\documentclass[conference]{IEEEtran}
\usepackage [utf8]{inputenc}

\ifCLASSINFOpdf
\else
\fi

\usepackage[most]{tcolorbox}

\newtcbtheorem{Summary}{\bfseries Summary}{enhanced,drop shadow={black!50!white},
  coltitle=black,
  top=0.3in,
  attach boxed title to top left=
  {xshift=1.5em,yshift=-\tcboxedtitleheight/2},
  boxed title style={size=small,colback=pink}
}{summary}

\newtcolorbox[auto counter]{Insight}[1][]{title={\bfseries Insight~\thetcbcounter},enhanced,drop shadow={black!50!white},
  coltitle=black,
  top=0.1in,
  attach boxed title to top left=
  {xshift=1.5em,yshift=-\tcboxedtitleheight/2},
  boxed title style={size=small,colback=pink},#1}

\usepackage{amsmath,amssymb,amsfonts}

\usepackage{bm}
\usepackage{cite}
\usepackage{bbding}
\usepackage{wasysym}
\usepackage{algorithm}
\usepackage[noend]{algpseudocode}
\usepackage{multirow}
\usepackage{mathrsfs}
\usepackage{balance}
 
\usepackage{url}
\usepackage{tikz}
\usepackage{savesym}
\savesymbol{iint}
\savesymbol{iiint}
\usepackage{amsmath}

\usepackage{hyperref}
\hypersetup{
	colorlinks=true,
	linkcolor=red,
	filecolor=magenta,      
	urlcolor=blue,
	citecolor=black
}

\usepackage[numbers, sort&compress]{natbib}

\usepackage{subcaption}

\usepackage{tabu}
\usepackage{threeparttable}
\usepackage{tabularx}

\usepackage{tabularray}

\usepackage{tikz}
\newcommand*\emptycirc[1][1ex]{\tikz\draw[thick] (0,0) circle (#1);} 
\newcommand*\halfcirc[1][1ex]{%
  \begin{tikzpicture}
  \draw[fill] (0,0)-- (90:#1) arc (90:270:#1) -- cycle ;
  \draw[thick] (0,0) circle (#1);
  \end{tikzpicture}}
\newcommand*\fullcirc[1][1ex]{\tikz\fill (0,0) circle (#1);}

\newcommand{\ec}{\emptycirc[0.8ex]}
\newcommand{\hc}{\halfcirc[0.8ex]}
\newcommand{\fc}{\fullcirc[0.9ex]}

\usepackage{bbding}

\usepackage{filecontents}

\usepackage{bbding}
\usepackage{wasysym}
\usepackage{booktabs}
\usepackage{longtable}
\usepackage{makecell}

\usepackage{bbm}

\usepackage{color}

\usepackage{xurl}

\usepackage{pythonhighlight}
\usepackage[scientific-notation=true,tight-spacing=true,
    round-mode          = places,
    round-precision     = 2,
    text-series-to-math = true ,
propagate-math-font = true
  ]{siunitx}

\begin{document}
%
\title{SafePowerGraph: Safety-aware Evaluation of Graph Neural Networks for Transmission Power Grids}

\author{\IEEEauthorblockN{Salah GHAMIZI}
\IEEEauthorblockA{LIST \thanks{LIST: Luxembourg Institute of Science and Technology} / RIKEN AIP\\
salah.ghamizi@list.lu}
\and
\IEEEauthorblockN{Aleksandar BOJCHEVSKI}
\IEEEauthorblockA{University of Cologne\\
a.bojchevski@uni-koeln.de}
\and
\IEEEauthorblockN{Aoxiang MA}
\IEEEauthorblockA{LIST\\
aoxiang.ma@list.lu}
\and
\IEEEauthorblockN{Jun CAO}
\IEEEauthorblockA{LIST\\
jun.cao@list.lu}
}

%


\IEEEoverridecommandlockouts
\makeatletter\def\@IEEEpubidpullup{6.5\baselineskip}\makeatother
\IEEEpubid{\parbox{\columnwidth}{
    Network and Distributed System Security (NDSS) Symposium 2024\\
    26 February - 1 March 2024, San Diego, CA, USA\\
    ISBN 1-891562-93-2\\
    https://dx.doi.org/10.14722/ndss.2024.23xxx\\
    www.ndss-symposium.org
}
\hspace{\columnsep}\makebox[\columnwidth]{}}

\maketitle

\begin{abstract}

Power grids are critical infrastructures of paramount importance to modern society and their rapid evolution and interconnections has heightened the complexity of power systems (PS) operations. Traditional methods for grid analysis struggle with the computational demands of large-scale RES and ES integration, prompting the adoption of machine learning (ML) techniques, particularly Graph Neural Networks (GNNs). GNNs have proven effective in solving the alternating current (AC) Power Flow (PF) and Optimal Power Flow (OPF) problems, crucial for operational planning. 
However, existing benchmarks and datasets completely ignore safety and robustness requirements in their evaluation and never consider realistic safety-critical scenarios that most impact the operations of the power grids.
We present SafePowerGraph, the first simulator-agnostic, safety-oriented framework and benchmark for GNNs in PS operations. SafePowerGraph integrates multiple PF and OPF simulators and assesses GNN performance under diverse scenarios, including energy price variations and power line outages. Our extensive experiments underscore the importance of self-supervised learning and graph attention architectures for GNN robustness. We provide at \url{https://github.com/yamizi/SafePowerGraph} our open-source repository, a comprehensive leaderboard, a dataset and model zoo and expect our framework to standardize and advance research in the critical field of GNN for power systems.

\end{abstract}

%

\section{Introduction}

\begin{figure*}[ht]
    \centering
    \includegraphics[width=0.8\linewidth]{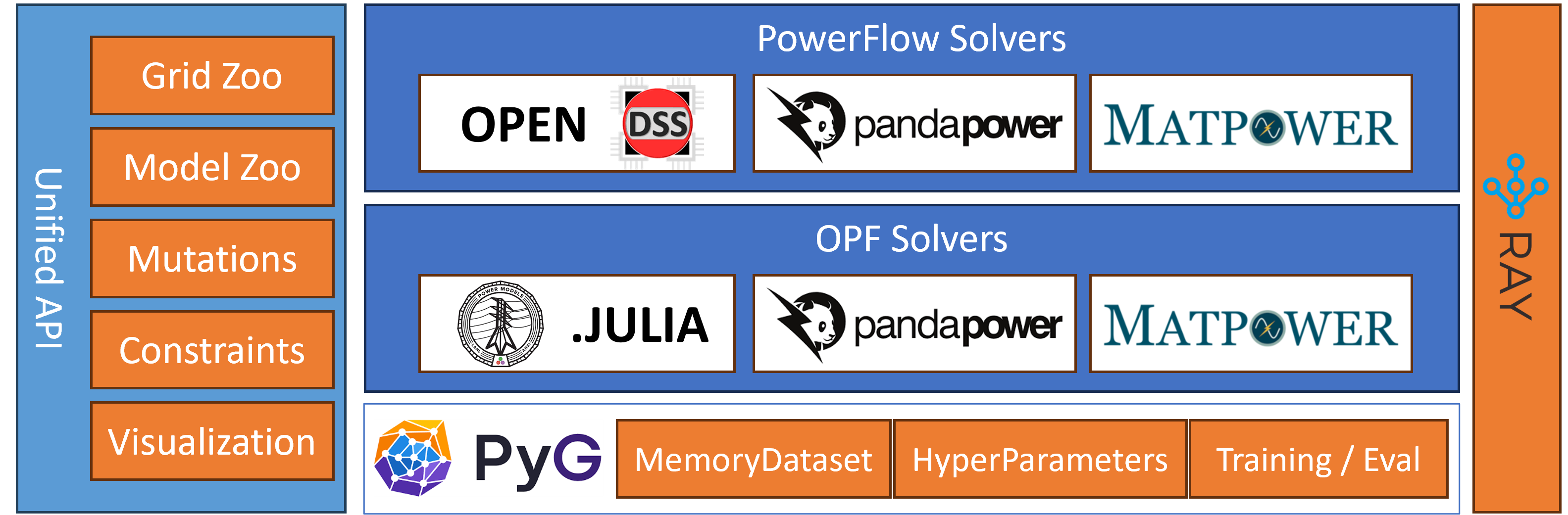}
    \caption{Architecture of the SafePowerGraph framework.}
    \label{fig:framework}
\end{figure*}

In recent decades, the energy sector has experienced significant transformations driven by a range of technological, economic, and environmental influences. One of the most remarkable trends is the development and maturity of renewable energy resources (RESs), and energy storage systems (ESs) that increase the complexity of monitoring and online-operations of power systems (PS).  

Transmission system operators (TSOs) require real-time tools for effective power systems operations, but current methods for grid analyses, hindered by their computational speed, cannot fully meet the challenges posed by RESs and ESs at large scale. 
Driven by these limitations, TSO are increasingly relying on machine learning (ML) techniques to support 
power operations. In particular, Graph Neural Networks (GNNs) have demonstrated their effectiveness in modeling the dynamics of the power grid and have emerged as the leading techniques for real-time solutions to solve optimizations of the alternating-current (AC) power system \cite{liao_review_2022,donon_neural_2020,owerko_optimal_2020}.

Solving AC power flow (PF) and AC Optimal Power Flow (OPF) problems is a routine task in power system operational planning. The PF problem involves determining the voltage magnitudes and phase angles of all buses in an electrical power system, ensuring that the power supplied meets the demand while satisfying the constraints of the network. 
Meanwhile, the OPF problem is an optimization problem. It seeks to find the best voltage settings in the system buses and generator power outputs to satisfy load requirements (energy demand), while satisfying network constraints and minimizing energy losses and energy generation costs. This optimization is not always scalable when considering large topologies, new components (RESs, ESs), and high variations in loads and energy prices.

GNNs are today de facto the predominant approaches to solve the PF problem \cite{gao_physics_2023,lopez-garcia_power_2023,ghamizi2024powerflowmultinet, lin_powerflownet_2023,jin_physics-informed_2024} and the OPF problem \cite{owerko_optimal_2020,song2023constraint,lopez-cardona_proximal_2022,liu_topology-aware_2023}. Few approaches are designed outside of pure academia and are implemented by TSO themselves \cite{donon_neural_2020, leyli2022lips}.  However, each publication and approach are evaluated under different experimental settings, using different PF and OPF solvers (we identified six) to generate their labeled datasets and rarely consider challenging and safety-critical evaluation scenarios with network perturbations. 

\begin{figure}
    \centering
    \includegraphics[clip, trim={0px 0px 0px 50px}, width=\linewidth]{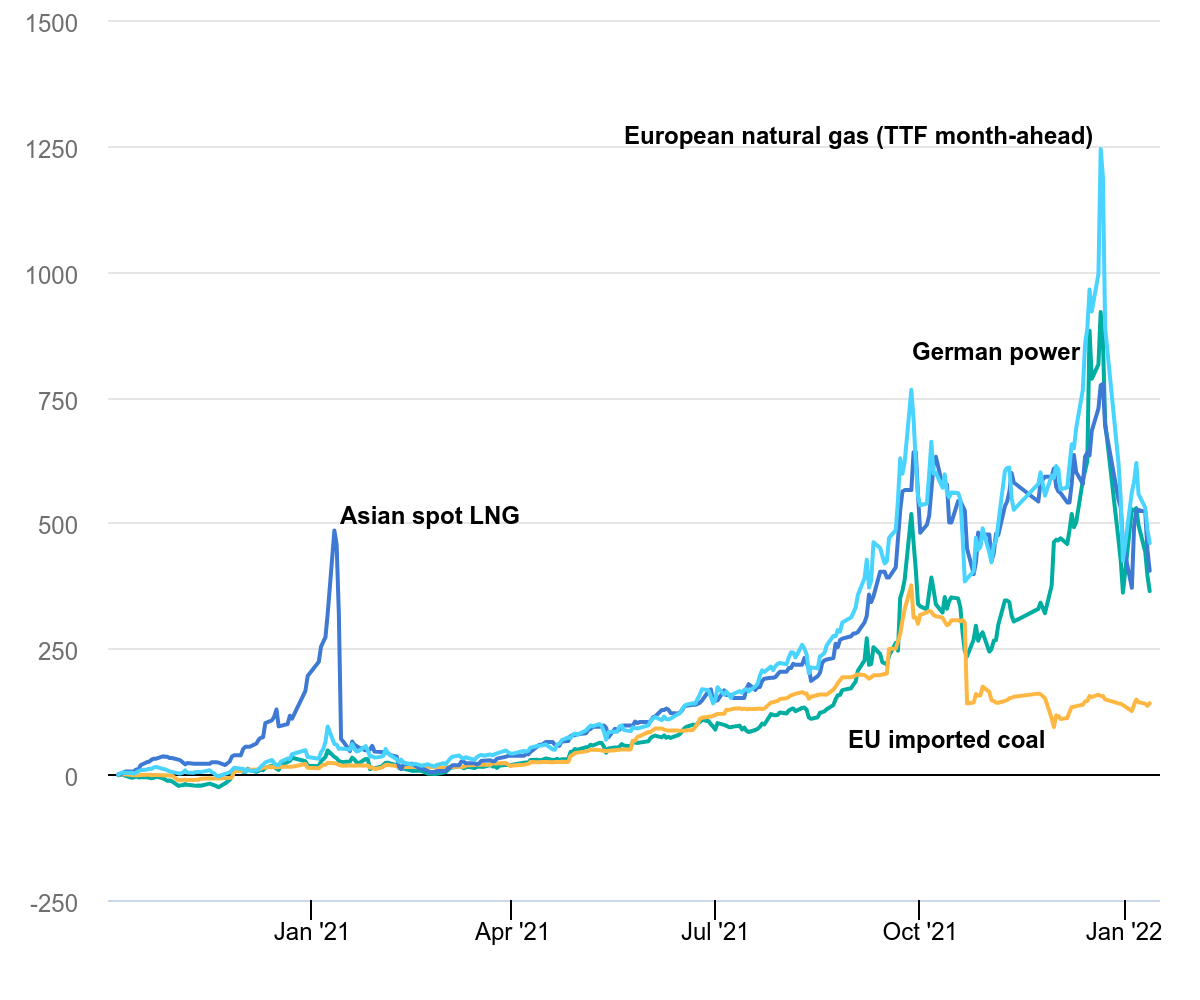}
    \caption{Evolution of energy prices, reported by the IEA \cite{iea2022}.}
    \label{fig:energy-prices}
\end{figure}

A comprehensive study on the perturbations of Nordic power grids \cite{entsoe2022} showed that these eight countries faced a total of 1972 major disturbances in 2022. These disturbances were associated with 1345 faults on overhead lines. Line faults and outages are among the main security threats to the stability and safety of power grids. Energy price surges are another real threat to the stability of the network, and the recent crises have challenged, for example, European energy prices (Fig. \ref{fig:energy-prices}). 

Public power grid datasets such as the Electricity Grid Simulated (EGS) dataset \cite{elgridsim}, the PSML \cite{PSML} dataset and the Simbench dataset \cite{simbench} are not specifically designed for machine learning applications on graphs and do not account for these realistic perturbations. To the best of our knowledge, there are only two benchmarks related to our work: LIPS\cite{leyli2022lips} and PowerGraph\cite{varbella_powergraph_2024}. We compare them in Table \ref{tab:related_works}. 

Their main limitations are that they (1) do not solve the variability problem of PF and OPF simulators in the community, and only support one simulation tool, (2) do not support safety-critical scenarios such as line outages and energy price variations, and (3) only consider traditional supervised learning (SL), while most of the new approaches are driven by self-supervised learning (SSL) with physics-informed ML \cite{liao_review_2022,huang_applications_2023}. These limitations make them unsuitable to unify and springboard future research in GNN for PS.  

We make up for these shortcomings with \textbf{SafePowerGraph}, the first safety-oriented framework and benchmark. Our framework is simulator-agnostic and natively integrates the four popular PF and OPF simulators, and provides safety assessment of GNN by considering supervised errors (MSE), and physical errors (constraints violations) for three settings: in-distribution (ID) scenarios, energy price variations scenarios, and power line outages scenarios. Our study demonstrates the relevance of self-supervised learning and Graph Attention architectures for the robustness of GNN for PS. 

In summary, we make the following contributions.

\begin{itemize}
    \item To the best of our knowledge, we are the first to define the problem of GNN vulnerability and safety for realistic PS operations. We introduce realistic scenarios and evaluation metrics tailored to this real-world critical application of GNN.
    
    \item We introduce an innovative integrated GNN framework, called \textbf{SafePowerGraph}, which is, to the best of our knowledge, the first practical framework and benchmark addressing the safey and robustness of GNNs within the PS operations context.

    \item We conduct extensive experiments on three standarized graph power grids networks and two critical safety edge-cases, and demonstrate the relevance of our benchmark. SafePowerGraph, uncovers the critical impact of GNN architectures and self-supervised learning to ensuring safe and robust GNNs for PS. SafePowerGraph is open-source and available at this repository: \url{https://github.com/[ANONYMIZED]}. 

    \item We share with the community a \textbf{Leaderboard} based on \textit{more than 200} evaluations to track the progress and the current state of the art in safety and robustness of graph deep learning models for PS, including SL and SSL approaches. The goal is to clearly identify the most successful ideas in GNN architectures and robust training mechanisms to accelerate progress in the field. Our leadearboard is accompanied with a \textbf{Dataset zoo} and a \textbf{Model Zoo} to standarize the research.

\end{itemize}

\section{Preliminaries}

Power grids consist of two elements: buses, which denote key points within the grid such as generation sites, load centers, and substations, and transmission or distribution lines that link these buses. Therefore, it is quite natural to visualize power grid networks as graphs, where buses and transmission lines are depicted as nodes and edges of the corresponding graph.

The buses are divided into three main categories: \textit{PV}, \textit{PQ}, and V$\theta$. \textit{PV} buses denote grid generators that supply and inject energy
. \textit{PQ} buses denote the grid's loads, which are components that consume energy (e.g. households). The V$\theta$ bus is referred to as a "slack bus" and acts as a reference point for grid operation. The state of each bus is defined by a set of variables depending on its type (Table \ref{tab:variables}).

Given the diversity of the components of the power grid, one can use a homogeneous graph representation where all buses have the same shape of features and labels, and their diversity is processed as homogeneous GNN with masking~\cite{lin_powerflownet_2023}, or leverage heterogeneous GNN to consider the node types separately~\cite{ghamizi2024hgnn}. 

Without loss of generalization, our work will focus on the heterogeneous undirected graph representation, but our framework based on Pytorch Geometric (PyG) supports all PyG graph representations. Although building GNN on top of homogeneous/heterogeneous directed/undirected graphs can lead to different properties, previous work \cite{ghamizi2024hgnn} showed the superiority of heterogeneous undirected graphs in terms of effectiveness and robustness.

\subsection{Graph Neural Networks}

Graph Neural Networks (GNNs) have emerged as powerful tools for learning on graph-structured data, which are prevalent in various domains such as social networks, molecular biology, and power grids. At their core, GNNs leverage the message-passing framework, where node features are iteratively updated by aggregating information from their neighbors. Early approaches, such as Graph Convolutional Networks (GCNs) \cite{KipfW17} applied convolutional operations to graphs, enabling the extraction of local patterns and capturing the graph structure effectively. Following this, numerous enhancements have been proposed to improve GNNs’ expressive power, such as GraphSAGE\cite{HamiltonYL17}, which addressed the scalability issues of GNNs by employing a sampling strategy to aggregate information from a fixed number of neighbors, thus allowing the model to handle large graphs efficiently.
Another significant development was Graph Attention Networks (GATs) \cite{VelickovicCCRLB18}, which leverages attention mechanisms to assign different weights to neighbor nodes, and DeeperGCN\cite{li2020deepergcn}, designed to mitigate vanishing gradient, over-smoothing and over-fitting when going deeper.

Recent advances include heterogeneous graph neural networks (HGNNs), which extend conventional graph neural networks (GNNs) to handle multiple types of nodes and edges effectively, for example, by leveraging metapath-based aggregation for richer representations \cite{sun2013mining, zhang2020hetero}.

GNNs support a wide range of tasks, broadly categorized into node- and graph-level tasks. Node-level tasks, such as node classification and node regression, focus on predicting the properties or labels of individual nodes within the graph. In power grids, predicting the state of specific buses belongs to this category. Graph-level tasks, on the other hand, involve predicting properties of entire graphs, such as molecular property prediction in chemistry or powerflow convergence in power grids. In addition, GNNs have been utilized for link prediction tasks, which aim to infer missing edges in partially observed graphs. In power systems, link prediction could consist of predicting vulnerable or non-robust lines of the grids, commonly referred to as ``contingency analysis". 

Our work focuses on two node-level regression tasks for power systems that are critical for power grid operations: Power Flow (PF) and Optimal Power Flow (OPF). 

\subsection{Power Grids Operations}

The scope of our work is to model and solve power flow and optimal power flow problems of transmission grids. We summarize in the following the equations of both the PF and the OPF problems using the polar form, and refer you to Appendix \ref{sec:app-A-powerflow} for detailed explanations. 

\vspace{3em}

\textbf{The Power flow problem}

The goal of the PF problem is to find the solution of the unknown variables in each bus given the known values of the buses and the state of the grid (topology and attributes of the grid), as shown in Table \ref{tab:variables}. $P^g_i, Q^g_i$ are the generated active and reactive power into bus $i$, $P^l_i, Q^l_i$ are the demand active and reactive power out of the bus $i$, $P_i, Q_i$ are the net real and reactive power injections with $P_i=P^g_i-P^l_i$ and $Q_i=Q^g_i-Q^l_i$, $V_i, \theta_i$ are the voltage magnitude and voltage angle at bus $i$.

\begin{table}
    \centering
     \caption{The variables that define the state of each bus in the PF and OPF problems.}
    \begin{tabular}{c|c|c|c}
    \toprule
       Problem & Bus type & Known variables (Input) & Unknown variables (target)\\
        \midrule
       & Slack bus (V$\theta$) & $V_i, \theta_i$ & $P_i, Q_i$\\
       PF & Load (PQ) & $P_i, Q_i$ & $V_i, \theta_i$ \\
        & Generator (PV) & $P_i, V_i$ & $Q_i, \theta_i$ \\
        \midrule
        
         OPF & & $P^l_i, Q^l_i$ & $P^g_i, Q^g_i, V_i, \theta_i $ \\
        \bottomrule
    \end{tabular}
    \label{tab:variables}
\end{table}

The model of the PF problem can be mathematically formulated using Kirchhoff's equations:\\
\begin{equation}
\label{eq:powerflow-base}
\resizebox{0.43\textwidth}{!}{$
    \begin{cases}
        P_i = V_i \sum_{k=1}^{N} V_k \left( G_{ik} \cos(\theta_i - \theta_k) + B_{ik} \sin(\theta_i - \theta_k) \right) \\
        Q_i = V_i \sum_{k=1}^{N} V_k \left( G_{ik} \sin(\theta_i - \theta_k) - B_{ik} \cos(\theta_i - \theta_k) \right)
    \end{cases}
$}
\end{equation}

for each bus $i, k \in \{1..N\}$ of the grid, where $G_{ik}$ and $B_{ik}$ denote two physical line properties, the conductance and susceptance, respectively.
Traditionally, PF problem is addressed with iterative solvers. The most popular use the Newton-Raphson method, linearizing power flow equations starting from an initial guess of the unknown variables in Table \ref{tab:variables}. We elaborate on the solver's optimization in Appendix \ref{sec:app-A-powerflow}.

We learn the PF problem by minimizing three losses:

\textbf{1- A supervised loss} that uses the solution from the solver (the oracle) as ground truth and minimizes the MSE between the predicted values of the GNN and the oracle's output.

\textbf{2- A self-supervised loss} that computes all the terms of equations \ref{eq:powerflow-base}, and minimizes the error between the left side and the right sides of the equations.  

\textbf{3- A boundary-violation loss} that models the limits of the components as soft constraints and pushes the outputs to satify them; e.g., to force the predicted active power of each generator within its active power capacity. 

\textbf{The Optimal Power Flow Problem}

The AC Optimal PowerFlow (OPF) problem is a more challenging problem than PF. It consists of finding the best solution to distribute electricity through power grids, considering factors such as demand, supply, storage, transmission limits, and costs, to ensure that everything runs smoothly and efficiently without overloading the system. The objective of this optimization is to fulfill the demand while \emph{minimizing the cost} of generation of various sources of energies.

The AC-OPF problem is formulated as follows:
\begin{align}
\min_{P^g_i, Q^g_i, V_i, \theta_i} \quad & \sum_{i=1}^{N_g} c_i(P^g_i)^2+b_i P^g_i+a_i \label{eq_cost} \\
\text{s.t.}~\forall i \quad & \resizebox{0.36\textwidth}{!}{$
    \begin{cases}
        P_i = V_i \sum_{k=1}^{N} V_k \left( G_{ik} \cos(\theta_i - \theta_k) + B_{ik} \sin(\theta_i - \theta_k) \right) \\
        Q_i = V_i \sum_{k=1}^{N} V_k \left( G_{ik} \sin(\theta_i - \theta_k) - B_{ik} \cos(\theta_i - \theta_k) \right)
    \end{cases}
$}
\end{align}
where $(c_i, b_i, a_i)$ represent the energy cost coefficients of the $i$th generator, $i \in \{1..N_g\}$. 

For our learning problem, we use the three aforementioned losses, where the oracle solves the AC-OPF problem. We also consider a regularization loss where the cost is also minimized:

\textbf{4- A cost loss} that models the cost objective. We use directly the absolute value of equation \ref{eq_cost} as regularization loss. 

Note that solving the OPF problem inherently solves the PF as well, by finding a unique solution for the cost objective.

\subsection{Robustness and Safety of Constrained NN}

Despite their effectiveness and scalability, GNNs face significant challenges in terms of robustness and generalization, and thus are not yet reliable to be deployed for critical applications \cite{liao_review_2022}. In particular, researchers have explored the lack of robustness of GNNs to out-of-distribution data \cite{gui_good_2022} and to adversarial perturbations \cite{jin2021adversarial}.

Early approaches to improve GNN robustness such as Graph Robustification \cite{ZugnerG19} and Adversarial Training \cite{madry2017towards} have successfully been explored to enhance GNNs’ resilience to such perturbations. 
Recent approaches \cite{jin2020graph,geisler2021robustness,gosch2024adversarial,wu2023adversarialweightperturbation} demonstrated the impact of architectures, training, and losses on the inherent robustness of GNN. However, few studies considered feasible perturbations under domain constraints. Geisler et al.~\citep{geisler2021generalization} explored the robustness of GNN-based solvers for the traveling salesman problem, where the perturbation space was restricted by the problem.

Constrained perturbation is a blooming field of research in computer vision and tabular machine learning. 
Ballet et al.~\cite{ballet2019imperceptible} focused on the importance of characteristics in the design of attacks, Mathov et al.~\cite{mathov2022not} took into account mutability, type, boundary, and data distribution constraints, and Simonetto et al.~\cite{simonetto2021unified} emphasized the importance of considering domain-constraints (inter-feature relationships) for crafting realistic perturbations.

Our work focuses on three realistic scenarios that GNNs for PS can encounter in the real world: load variations, price variations, and line outages. We generate valid perturbations by ensuring that all the domain constraints of the grid are satisfied, in particular the power flow equations defined in Eq. \ref{eq:powerflow-base}, and the boundaries of the features (Appendix \ref{sec:app-B-features}).

\begin{table*}[t]
\centering
\footnotesize
\renewcommand\arraystretch{1.3}
\caption{Comparison between our framework SafePowerGraph and GNN related works. \ec \ indicates ``Not Considered", \hc \ indicates ``Partially Considered", and \fc \ indicates ``Fully Considered". PP stands for ``PandaPower", OD for ``OpenDSS", PW for ``PowerModels", MP for ``MatPower", LSG for ``LightSim2Grid".}
\label{tab:related_works}
\begin{threeparttable}
\begin{tabularx}{\textwidth}{c|c|c|*{3}{>{\centering\arraybackslash}X}}
\hline
\multicolumn{1}{c|}{\multirow{2}{*}{\textbf{Features}}} &
  \multicolumn{1}{c|}{\cite{gao_physics_2023,lopez-garcia_power_2023,ghamizi2024powerflowmultinet}} &
  \multicolumn{1}{c|}{\cite{song2023constraint,lopez-cardona_proximal_2022}} &
  \multicolumn{3}{c}{\textbf{Benchmarks}} \\ \cline{4-6} 
 &
 \multicolumn{1}{c|}{\cite{lin_powerflownet_2023,jin_physics-informed_2024}} &
  \multicolumn{1}{c|}{\cite{liu_topology-aware_2023}} &
  \multicolumn{1}{c|}{\textbf{LIPS}\cite{leyli2022lips}} &
  \multicolumn{1}{c|}{\textbf{PowerGraph} \cite{varbella_powergraph_2024}} &
  \multicolumn{1}{c}{\textbf{OURS}} \\
\hline
Simulators & MP \emph{OR} PP \emph{OR} OD & PW \emph{OR} MP & LSG & MP & PP+OD+PW+MP \\
\hline
Tasks & PowerFlow & OPF & PowerFlow & PowerFlow+OPF & PowerFlow+OPF  \\
\hline
Load variation & \hc & \hc & \hc & \hc & \fc \\
Price variation  & \ec & \ec & \ec & \ec & \fc  \\
Line outage  & \hc & \ec & \fc & \ec & \fc  \\

\hline
Architectures & \hc & \hc & \hc & \fc & \fc \\
Safety constraints & \hc & \hc & \fc & \ec & \fc  \\

\hline
Sup Learning & \hc & \hc & \fc & \fc & \fc \\
Self-Sup Learning & \hc & \hc & \ec & \ec & \fc  \\
\hline
\end{tabularx}
\end{threeparttable}
\end{table*}
\section{Problem Formulation}

PF and OPF are node-level regression tasks, and the target (the ground truth of the state) for each node is a continuous vector provided by an oracle. At each perturbation, both the prediction and the target will then vary. 

Contrary to traditional robustness assessment, where the target is not supposed to vary under small perturbations, the robustness of power grids is measured based on an acceptable threshold of error compared to the oracle. The targets of the nodes require expensive solvers to obtain the new ground truths. 
We formulate in the following the problem of heterogeneous GNN robustness under constrained perturbations.

We consider a heterogeneous graph, denoted as $\mathcal{G}=(\mathcal{V},\mathcal{E},\mathbf{M})$, that consists of a node set $\mathcal{V}$ and a link set $\mathcal{E}$ and an adjacency matrix $\mathbf{M}$. 
A heterogeneous graph is also associated with a node-type mapping function $\phi:\mathcal{V}\rightarrow \mathcal{A}$ and a link-type mapping function $\psi: \mathcal{E}\rightarrow \mathcal{R}$. $\mathcal{A}$ and $\mathcal{R}$ denote the sets of predefined node types and link types, where $|\mathcal{A}|+|\mathcal{R}|>2$.

For both PF and OPF problems, we consider a homogeneous link type $|\mathcal{R}|=1$, and the following node types: $\mathcal{A} = \{\text{bus, line, transformer, generator, slack, load, capacitor} \}$. 

We denote by $n_i$ the number of nodes of type $i$ in the graph $\mathcal{G}$, by $f_i$ the number of features of nodes of type $i$, and by $t_i$ the number of outputs of nodes of type $i$. 
Each node of type $i$ has a set of features $X_i = \{X_i^k ~/~ \forall k \in 1..f_i\}$. The details of the features and outputs of each node are in Appendix \ref{sec:app-B-features}.
 
We denote by $\mathcal{X}=\{X_i \in \mathbb{R}^{n_i \times f_i}$ for $i \in \mathcal{A} \}$ an input example, defined as the heterogeneous dictionnary of features of the nodes of the graph $\mathcal{G}$, and denote by $\mathcal{Y}=\{y_i \in \mathbb{R}^{n_i \times t_i}$ for $i \in \mathcal{A} \}$ its correct target. 
Let $h: (\mathcal{X},\mathbf{M})  \rightarrow \mathcal{Y}$ be a graph neural network that outputs a regression value $h=\{h_i \in \mathbb{R}^{n_i \times t_i}$ for $i \in \mathcal{A} \}$, and $g:  (\mathcal{X},\mathbf{M})  \rightarrow \mathcal{\hat{Y}}$ an oracle that outputs the ground truth value using traditional exact solvers (e.g., Newton Raphson). 

Let $\Delta$ be the space of allowed perturbations. 
We define a $(\Delta, \mu)$-robust model $h$ to be any model (e.g. GNN) that satisfies 
\begin{equation}
\label{eq:solver}
\forall ~\delta \in \Delta : ||h(\mathcal{X}+\delta) - g(\mathcal{X}+\delta))||_p \leq \mu,
\end{equation}
where $||\cdot||_p$ is a defined distance (e.g. an $L_p$-normed distance) and $\mu$ is an acceptable error threshold.

In image classification, the set $\Delta$ is typically chosen as the perturbations within some $L_p$-ball around $\mathcal{X}$, that is, $\Delta_p = \{ \delta \in \mathbb{R}^d, ||\delta||_p \leq \epsilon\}$ for a maximum perturbation threshold $\epsilon$.

Power grid data are by nature different from images. 
Each feature $x_{ij} \in X_i$ of the graph represents a physical property (e.g., line resistances, length, etc.) or a smart meter measurement (e.g. load's active power). Thus, $x_{ij}$  has to respect physical constraints to be valid. 

A minimal requirement is to respect the boundary constraints of each feature, and a sufficient requirement is that the perfect solver converges to a solution when optimizing the perturbed Pf or OPF problem. In practice, traditional solvers are not perfect and are dependent on their hyper-parameters (threshold, budget, ...), however, we assume that our oracle solver is perfect.  
A robust GNN is then defined as follows:
\begin{equation}
\label{eq:solver}
\forall \delta~ g(\mathcal{X}+\delta) \text{~converges~}\rightarrow{} ||h(\mathcal{X}+\delta) - g(\mathcal{X}+\delta))||_p \leq \mu
\end{equation}

\paragraph{Threat Model} 
Within the Power Grid operations problems outlined above, a critical concern is the threat of unexpected grid variations during model inference,
which compromises the safety of the grids and the integrity of operations of the Transmission System Operator (TSO). To scope our
discussion, we assume that the variations are benign, that is, they have not been crafted specifically to hurt the ML system but represent \emph{black swans} events with major impacts on the grid.



\paragraph{Design Requirements}

This paper aims to develop a framework to assess and mitigate the impact of Power Grid perturbations in the context of GNN-based powerflow and optimal power flow problems.
Our method should be suitable for the Power Grid scenario that realistically poses major threats to the grid's optimal operation.
Specifically, our design requirements are as follows.

\textbf{P1 - Energy price variations:} Each generator $i$ has a polynomial generation cost function as follows: $c(P^g_i) = c_i+ b_i \cdot P^g_i + c_i \cdot (P^g_i)^2$, where $c_0$ is the offset active power cost, $c_1$ the linear cost per Mvar and $c_2$ the quadratic cost per MW. For each studied grid, we only vary the linear cost $c_1$, and never consider switching off completely a generator. 

The cost parameters only impact the optimization of the OPF problem and are ignored for the PF problem.

\textbf{P2 - Load variations:} Each consumer is defined by its load active power $P_l$ and reactive power $Q_l$. While the training data uses a time series of real world consumptions, a robust model should generalize to extreme variations of loads for sub-parts or the entirety of the graph. We model these variation with a multipliper sampled from a uniform distribution. The distribution of this multiplier and the percentage of affected loads within the the grid reflect the severity of the perturbation. 

\textbf{P3 - Line outage:} A line outage refers to the loss or failure of a transmission line within an electrical power system. Transmission lines are vital for carrying electricity from power plants to substations and, eventually, to consumers. When a line goes out of service due to a fault, maintenance, or any other reason, it can have several starting with the redistribution of loads that requires power flow change, to possible cascading failures leading up to widespread blackouts. To simulate outages, we evaluate the N-1 reliability criterion. It states that the power system must be capable of withstanding the failure of any single component without causing widespread outages. We randomly remove one line of the grid during evaluation and assess the robustness of the GNN on this degraded grid.
\section{SafePowerGraph}

\begin{figure*}[!t]
    \centering
    \includegraphics[width=\linewidth]{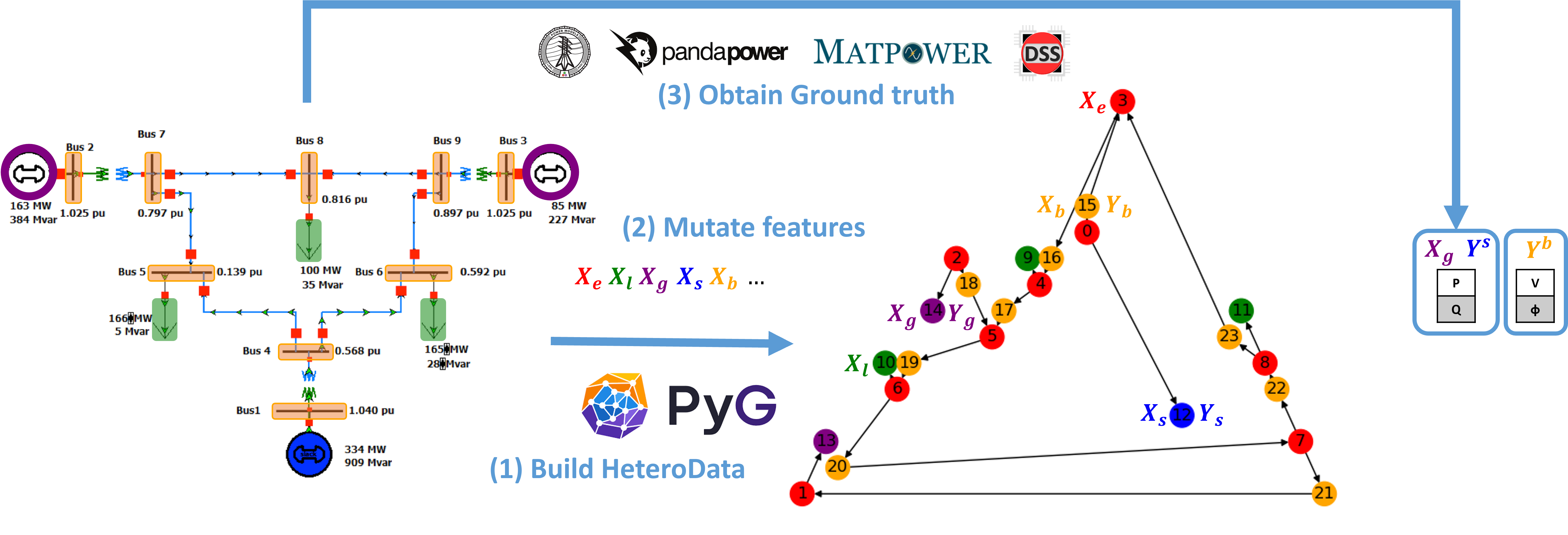}
    \caption{Embedding for a 9-bus grid. In purple the generators, blue the slack, orange the buses, green the loads, and red the lines.}
    \label{fig:embedding}
\end{figure*}

We introduce the architecture of our framework in Fig. \ref{fig:framework}. Our Python API abstracts both Power Systems (PS) and Machine Learning (ML) functionalities for practitioners from both fields. For ML practitioners, SafePowerGraph is a standarized datasets and models zoo with a unified benchmark to evaluate node-level GNN tasks on complex graphs: Constrained heterogeneous graphs modeling physical equations and a nonlinear minimization problem. For PS practioners, SafePowerGraph is a GNN-powered OPF and PF simulator that supports major PS formats and tools, and achieves precise, robust, and safe predictions. The main features of our API are:

\subsection{PowerFlow and OPF Simulation}

Our framework supports the most popular Power Grid libraries for full interoperability:

\paragraph{PandaPower (PF, OPF)} PandaPower \cite{pandapower.2018} is an open-source Python library that provides tools for the modeling, analysis, and optimization of electrical power systems. It is built on top of the popular data analysis library pandas, PandaPower and the power flow solver PYPOWER.
\paragraph{MATPOWER (PF, OPF)} MATPOWER \cite{matpower.2011} is an open-source MATLAB package that provides tools for steady-state power system simulations and optimizations. It supports large-scale PF analysis and OPF optimizations.
\paragraph{OpenDSS (PF)} OpenDSS (Open Distribution System Simulator) \cite{opendss} is a popular open-source software for simulating electric power distribution systems. 
In particular, it natively supports unbalanced distribution grid power flow estimations and is highly scalable.
\paragraph{PowerModels (OPF)} PowerModels \cite{powermodels} is a Julia/JuMP-based open-source package designed for solving complex power network optimization problems such as OPF, unit commitment and transmission expansion planning. It is highly extensible and scalable to larger grids.

\subsection{Graph Data Generation}

In Fig. \ref{fig:embedding} we summarize how each graph is generated from a power grid.
Starting from an initial grid descriptor (in MatPower, PandaPower or PyPower formats), we generate a Pytorch HeteroData with each component as a distinct subgraph (Step (1)). Each node can be of type: bus, load, generator, slack, line, transformer, or capacitor, and is associated with its distinct set of features $X_b$, $X_l$, $X_g$, $X_s$, $X_e$, $X_t$, $X_c$ respectively. These features support perturbations dependant of their types. For example, generators and slack nodes support the mutation of the price features (i.e., the cost of each unit generated) and their maximum generation capacity (Step (2)). 

Each node is also associated with a set of constraints. They can be included during training as regularization losses, or enforced with clamping (e.g. boundary constraints).  All features, mutations, and boundary constraints are detailed in the Appendix \ref{sec:app-B-features}.

For each mutated grid, we run in Step (3) a simulation and solver to obtain the ground truth of the PF or OPF optimization. This solution consists of the active and reactive power of each generator and the slack nodes ($Y_g$, $Y_s$) and the voltage magnitude and angle for each bus node ($Y_b$). 

Building large graph datasets across multiple mutations can be expensive. We distribute our datasets in a zoo following the Open Graph Benchmark (OGB) taxonomy \cite{hu2020open}.

\subsection{Training}

\begin{figure}[htbp]
  \centering
   \includegraphics[width=\linewidth]{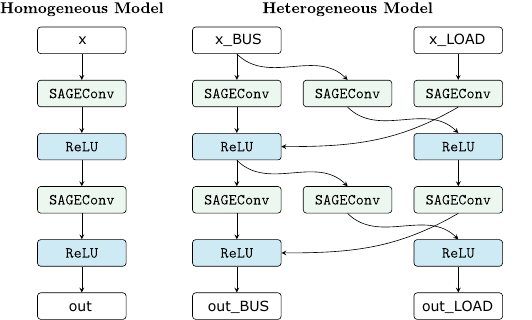} 
  \caption{Transforming a homogeneous GNN to a heterogeneous GNN by combining message passing across multiple types. Here with two node types \textbf{bus} and \textbf{load}, and Sage layers.}
  \label{fig:hgnn-graphsage}
\end{figure}

Heterogeneous graph training supports multiple existing architectures, including GCN, Sage, and GAT. It consists of interleaving the message functions on each node type individually. We present in figure \ref{fig:hgnn-graphsage} the extension of a standard single-node type architecture to two node types (bus and load).

Each of our models consists of a succession of heterogeneous graph layers (the backbone), and a set of fully connected layers that outputs the final predictions (head) \cite{liu_topology-aware_2023}. Following \cite{ghamizi2024hgnn}, we extend this standard structure by computing the loss of each node type separately. The splitted loss computations 
allows to dynamically update the weight (the contribution) of each node type's loss in the backpropagated loss. 
SafePowerGraph supports weighting strategies because previous research demonstrated that adequate weighting can significantly improve the robustness of multitask models \cite{ghamizi2022adversarial,ghamizi2023gat}. 

\textbf{Training loss:} We incorporate the constraints of the problem in the loss of our GNN as regularization terms.
We transform each of the constraints and optimization objectives into a regularization loss.
Given a training graph $\mathcal{G}$, its set of node types $\mathcal{A}$, its features $\mathcal{X}$, its associated ground truth prediction $\mathcal{Y}$ and its predicted output $y$, the training loss function of our GNN (parametrized by $\Theta$) becomes:

\begin{align}
    \mathcal{L}(\Theta) &:= \lambda_{b} \|y_b - \mathcal{Y}_b \|^2_2 + \lambda_{s} \|y_s - \mathcal{Y}_s \|^2_2 + \lambda_{g} \|y_g - \mathcal{Y}_g \|^2_2\nonumber \\
    &\quad + \lambda_{b,i} \sum_{i \in \mathcal{A}} \Bigg[\sum_{\omega \in \Omega_b} \text{ctrloss}(x_i, y_i, \omega) \nonumber \Bigg] \\
    &\quad + \lambda_{s} \cdot \text{sslloss}(x, y) \nonumber \\
    &\quad + \lambda_{c} \cdot \text{costloss}(x_g, y_g,x_s, y_s) 
    \label{eq:losses}
\end{align}

where the first term is the supervised loss over the bus, slack, and generator outputs.
The second term captures the weighted constraints violations of the power grid given the set of constraints $\Omega$ of each type of node.
The third term is the self-supervised loss that solves the equations of the power grid in \ref{eq:powerflow-base}.
The last term is the energy cost minimization loss considering the prices of the generators and the slack node.

Our study in 
Appendix \ref{sec:app-C-cost} confirms that optimal solutions in our scenarios require all the losses.
Our pretrained models can be downloaded with our API and the results are available on a public, easy-to-use benchmark available on \url{https://github.com/yamizi/SafePowerGraph}.

\subsection{Quantitative Evaluations}

In our evaluation, by default, we report three errors: 

\textbf{Supervised error.} We provide the normalized square error for predicted variables for each type of nodes averaged across all the nodes of this type per grid. For example, the reported MSE for generators in a 9-bus grid is averaged over the two generators of this grid. The final results is the average self-supervised error across all the validation grids. 

\textbf{Self-supervised error.} We report the violations to the powerflow equations as an MSE over all the buses of the grid. The final results is the average self-supervised error across all the validation grids. 

\textbf{Boundary errors.} Given a violation threshold, for each node we consider all the boundary violations as binary. A graph is considered 'valid' if it has no violation. The final results is the percentage of invalid graphs in amonth the validation graphs.

\subsection{Qualitative Evaluations}
Our framework supports an in-depth analysis of the grid and the behavior of individual lines, buses, and constraints.

\textbf{Constraints Evaluation:} The framework records individual violations for powerflow and boundary constraints. They are stored in dataframe format for individual visualizations and analyses of the buses, the generators, the slack nodes.

\textbf{Loading Visualization:} The framework allows the visualization of the the bus and line loading for solutions computed by a mathematical solver and by GNN. Buses are expected to have voltages under 1.0 per unit (pu) ideally, and at worst 1.1 pu. Line loading is ideally under 80\%, and dangerous over 100\%. Overloaded lines can lead to outages and have cascading effects. Our framework record these metrics and provides a simple visualisations by leveraging Pandapower \cite{pandapower.2018} plotting API for all simulators and GNN.  

\subsection{Other components}

SafePowerGraph integrates RAY parallelization to mutate and generate embeddings for thousands of graphs seamlessly. The calls to solvers are also parallelized in CPU and the training leverages GPU parallelization. The framework supports natively experiment trackers such as Comet and WandB.

\section{Benchmarking GNN for Power Grids}
\label{sec:results}
We focus the empirical study on the harder OPF problem, and defer some results for the PF problem to the Appendix \ref{sec:app-B}.

\subsection{Experimental Settings}

\textbf{Power grid topologies:}
We evaluate three commonly used topologies: WSCC 9-Bus system, IEEE 30-Bus system, and IEEE 118-Bus system.
The first case represents a simple system with 9 buses, 3 generators, and 3 loads.
The IEEE test cases represent an approximation of the American Electric Power system. The IEEE 30-Bus system has 30 buses, 5 generators, and 24 loads, and IEEE 118-bus has 118 buses, 19 generators, 35 synchronous condensers, 177 lines, 9 transformers, and 91 loads.

\textbf{Power grid mutations:}
For each experiment, we generate 1000 valid mutants (that is, for which the solver converged to a solution). We generate 800 training graphs using load variations from a real-world timeseries, and 200 test graphs are generated following different perturbation scenarios. The initial scenario is an In-Distribution mutation, where the loads are mutated following the same distribution of the training set (while ensuring no data leakage between train and test). We refer to this scenario as \textbf{ID}. For the second scenario (\textbf{line outage}), we randomly disconnect one line from the grid. In the last scenario, we randomly mutate the price of production of individual generator within the original generator price boundaries. We refer to this last scenario as \textbf{Price variation}. 

\textbf{GNN architectures:}
We evaluated three GNN layer architectures: Graph Convolution (cgn), SageConv (sage), and Graph Attention (gat). We use two layers and run a hyperparameter search on the number of features and hyperparameters of the layers (with 20\% cross-validation). In evaluation we report the best performing model for each architecture. 

After the two graph layers, each model is composed of two fully connected layers with 128 features each. The output of the models depends on the size of the grid and is of size $2 \times (n_{bus} +n_{generators} + 1)$ to learn a regression task for the OPF solutions.

\textbf{Evaluation Metrics}
For the quantitative study, we report the supervised normalized MSE for the active and reactive powers of the generators and the slack node, and for the voltage magnitude and angle of the buses. We refer to them as $P_{gen}$, $Q_{gen}$, $P_{slack}$, $Q_{slack}$, and $V$, $\theta$, respectively. We run our evaluations over three random seeds, report/plot in the main paper the mean values, and report in the appendices the standard deviations.

For the qualitative study, we evaluated the boundary constraints with a tolerance of $10^{-4}$ (as they are hard physical constraints, such as the generation capacity), and we used a tolerance of $10^{-2}$ for the power flow equations. 

\textbf{Training and optimization:}
We trained all models with the 4x32Gb V100 GPU, for 1000 epochs and batch size of 128. We used Adam optimizer and a multistep learning rate, starting at 0.001 and decaying by 0.5 at epochs $\in \{500,750,875\}$. 

We evaluated three weighting strategies to weight $\lambda_i$ in the loss equation \ref{eq:losses}. A strategy with equal weights, a strategy with softmax randomized weights at each batch and epoch, and a normalized weight based on the cardinality of each loss term (number of buses, generators, constraints,...) We report an empirical study on the best weighting strategies in Appendix \ref{sec:app-C-weighting}.
 
\begin{figure*}[h]
\centering
\begin{subfigure}{0.3\linewidth}
    \includegraphics[clip, width=\textwidth]{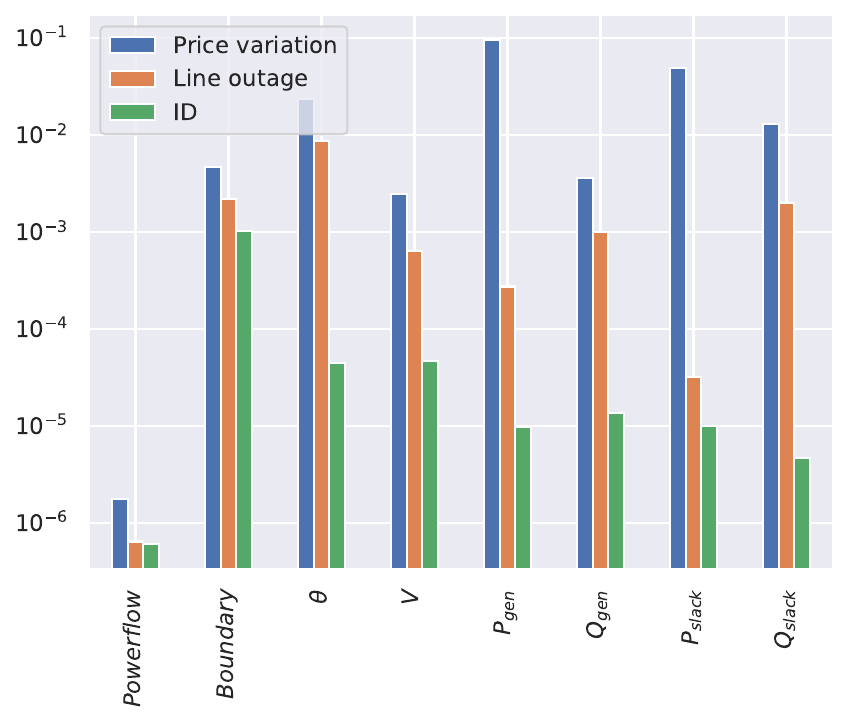}
    \caption{9-bus}
    \label{fig:variations-case9}
\end{subfigure}
\begin{subfigure}{0.3\linewidth}
    \includegraphics[clip, width=\textwidth]{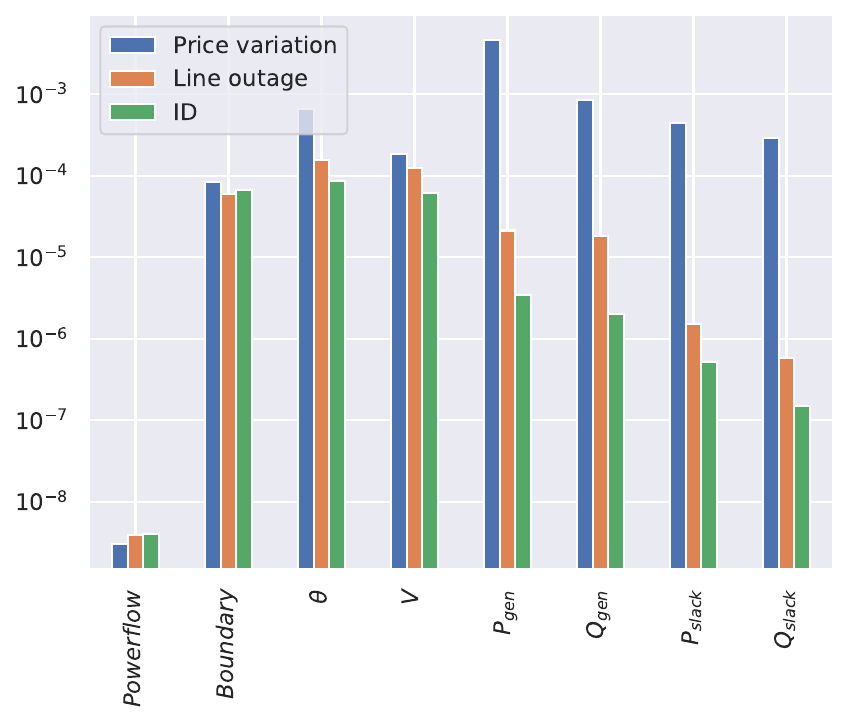}
    \caption{30-bus}
    \label{fig:variations-case30}
\end{subfigure}
\begin{subfigure}{0.3\linewidth}
    \includegraphics[clip,width=\textwidth]{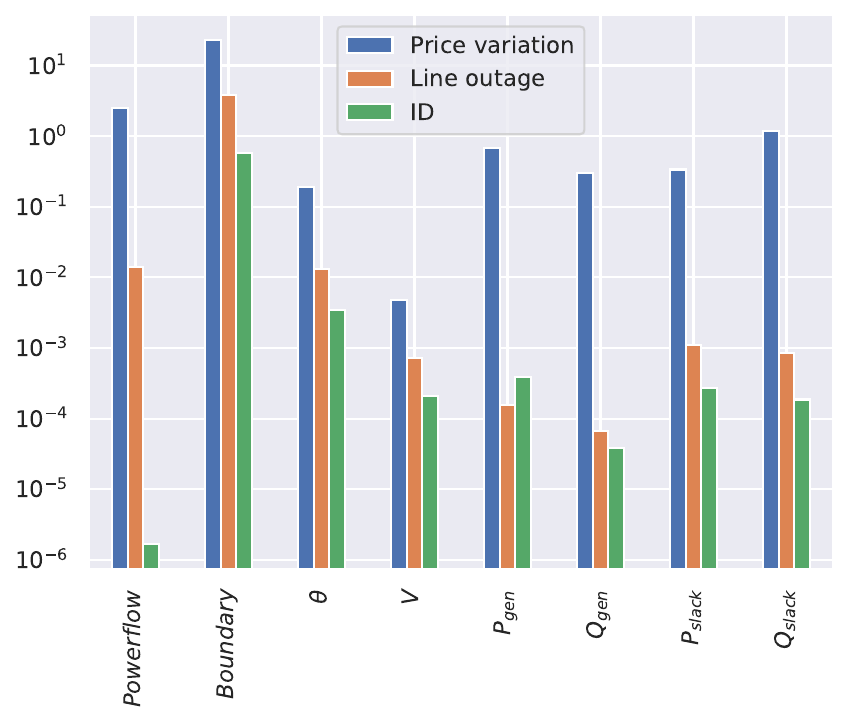}
    \caption{118-bus}
    \label{fig:variations-case118}
\end{subfigure}
\hfill

\caption{Impact of power grid perturbations on the robustness and safety of the OPF predictions.}
\label{fig:variations}
\end{figure*}

\subsection{OPF quantitative evaluation}

\textbf{Line outages.} We report in orange in Fig. \ref{fig:variations} the impact of line outages.
They have limited impact on the errors of the Powerflow equations compared to the ID setting for 9-bus and 30-bus. The error varies from $\num{6.015817e-07}$ to $\num{6.419584e-07}$ in 9-bus and from $\num{4.049878e-09}$ to $\num{3.838561e-09}$ in 30-bus. 

Line outages, however, significantly increase the Powerflow equation errors for the larger 118-bus from $\num{0.000002}$ to $\num{0.013825}$. Line outages have limited impact on the boundary violation errors across all the grids sizes. 
The remaining metrics show a moderate increase in error between the ID and line outage scenarios across all grids. 

\begin{Insight}[boxed title style={colback=pink}]
GNN models are robust to N-1 line outages on small grids. However, in larger grids, the predicted solutions increase the powerflow grid errors by a 4-order of magnitude and the optimal solution errors up to 100 times.
\end{Insight}

\textbf{Impact of price variations.} We report in blue in Fig. \ref{fig:variations} the impact of price variations.
There is a significant error increase in the GNNs predictions across all grid sizes. That is, the error of the reactive power of the slack node increases from $\num{4.668193e-06}$ to $\num{0.012945}$ the 9-bus grid, from $\num{1.488563e-07}$ to $\num{2.939446e-04}$ for the 30-bus grid and from $\num{0.000187}$ to $\num{1.191636}$ for the 118-bus grid. Meanwhile, the powerflow error is slightly affected for small grids, but increases significantly in the 118-bus grid, from $\num{0.000002}$ to $\num{2.517417}$.

\begin{Insight}[boxed title style={colback=pink}]
GNN models are not at all robust to price variations across all the grid sizes, and the bus powerflow errors increase up to 6-order of magnitude for the larger grids.
\end{Insight}

\begin{figure*}[ht]
\centering
\begin{subfigure}{0.3\linewidth}
    \includegraphics[clip, width=\textwidth]{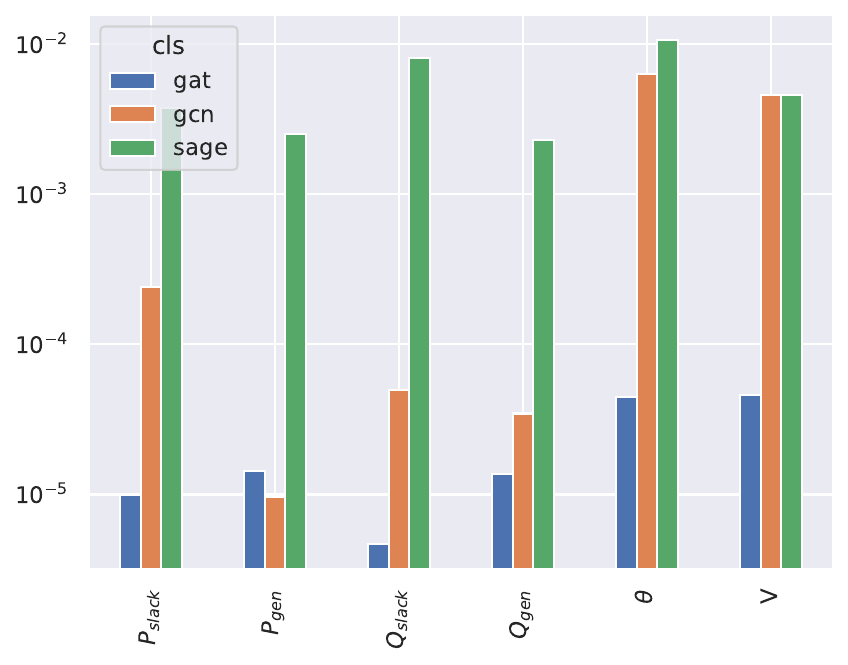}
    \caption{9-bus with ID variations}
    \label{fig:arch-case9-load}
\end{subfigure}
\begin{subfigure}{0.3\linewidth}
    \includegraphics[clip, width=\textwidth]{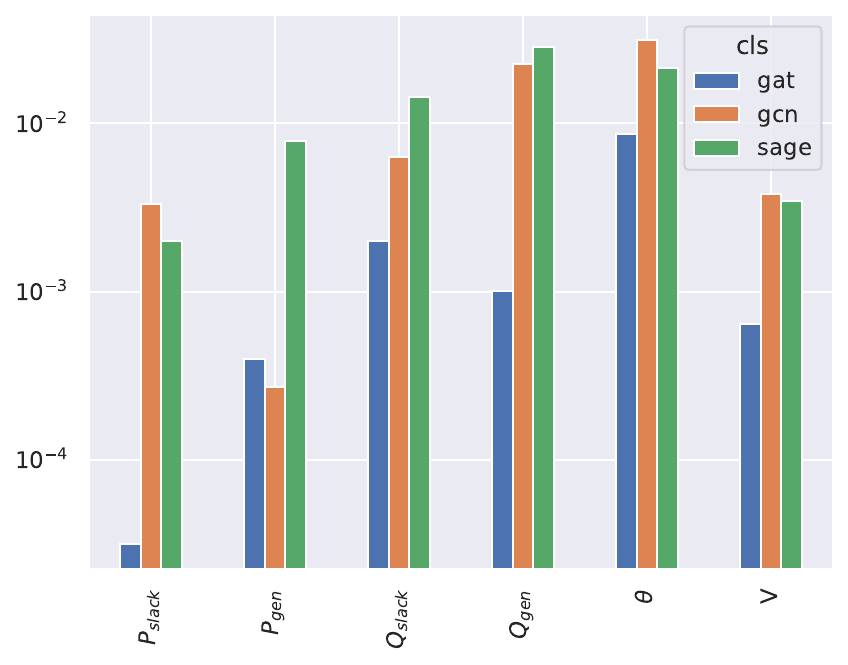}
    \caption{9-bus with line outages}
    \label{fig:arch-case9-line}
\end{subfigure}
\begin{subfigure}{0.3\linewidth}
    \includegraphics[clip, width=\textwidth]{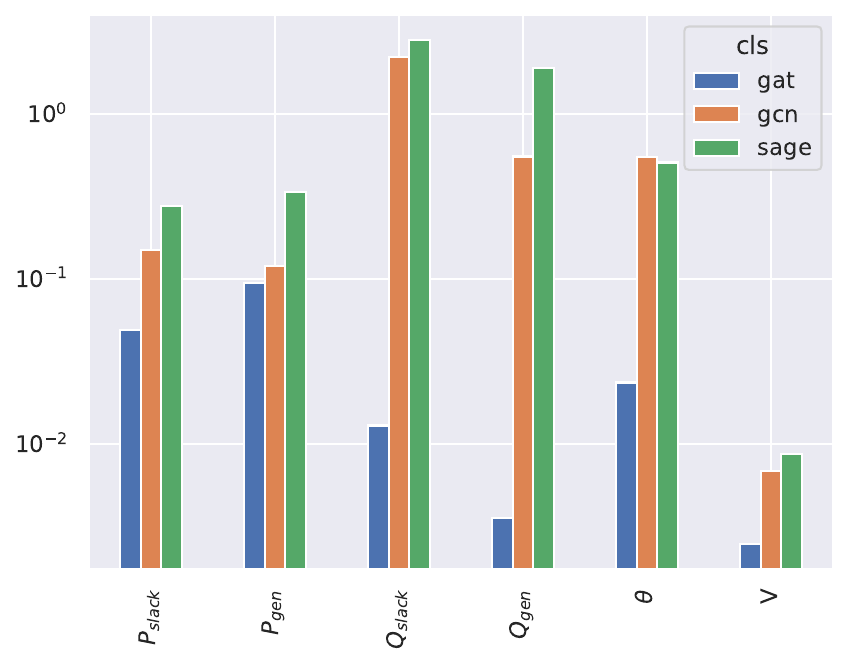}
    \caption{9-bus with price variations}
    \label{fig:arch-case9-cost}
\end{subfigure}
\hfill
\begin{subfigure}{0.3\linewidth}
    \includegraphics[clip, width=\textwidth]{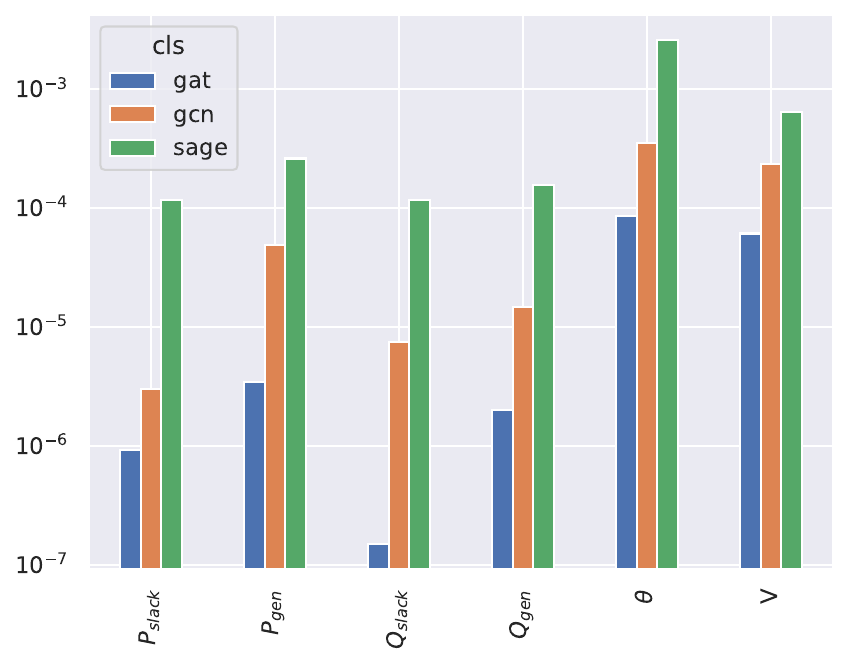}
    \caption{30-bus with ID variations}
    \label{fig:arch-case30-load}
\end{subfigure}
\begin{subfigure}{0.3\linewidth}
    \includegraphics[clip,width=\textwidth]{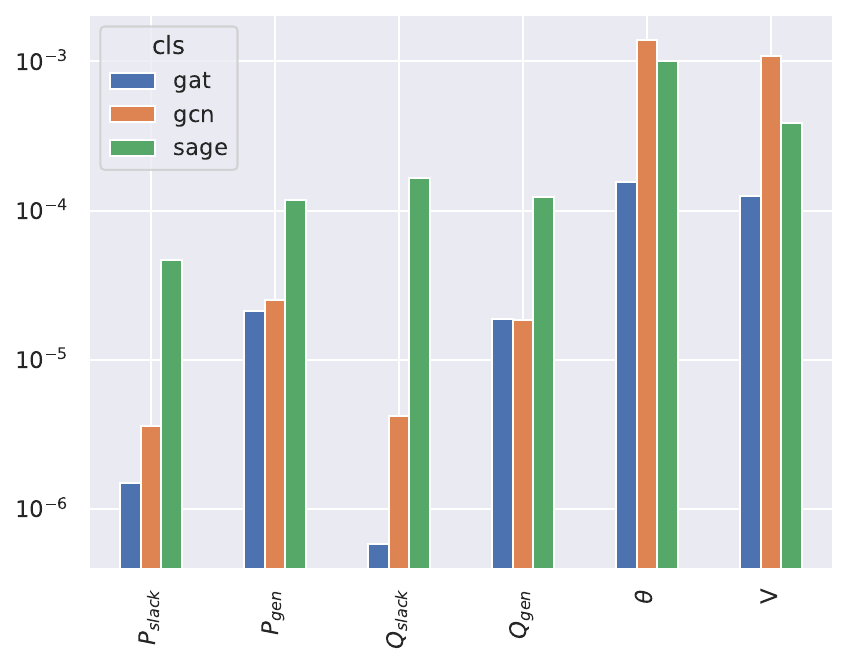}
    \caption{30-bus with line outages}
    \label{fig:arch-case30-line}
\end{subfigure}
\begin{subfigure}{0.3\linewidth}
    \includegraphics[clip,width=\textwidth]{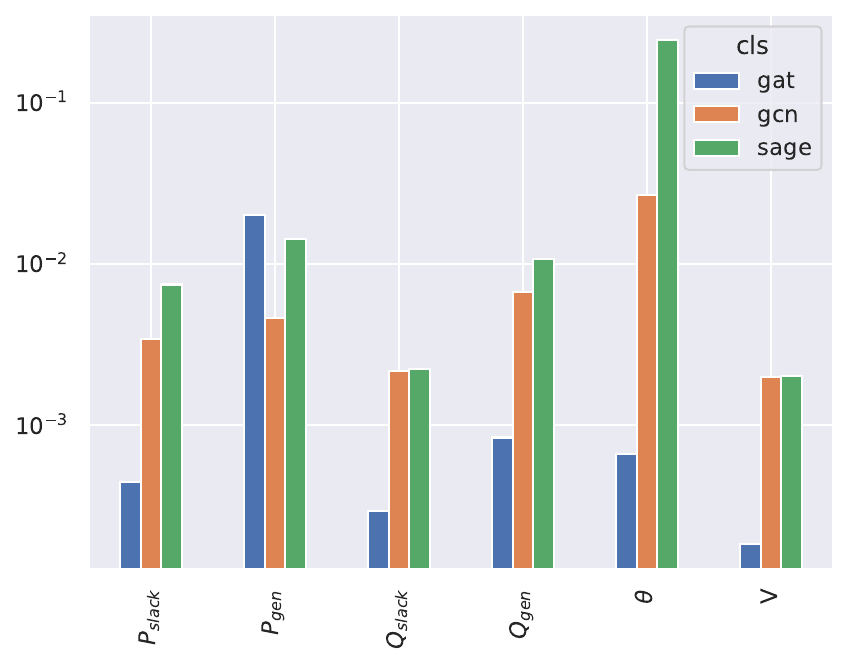}
    \caption{30-bus with price variations}
    \label{fig:arch-case30-cost}
\end{subfigure}
\hfill
\begin{subfigure}{0.3\linewidth}
    \includegraphics[clip, width=\textwidth]{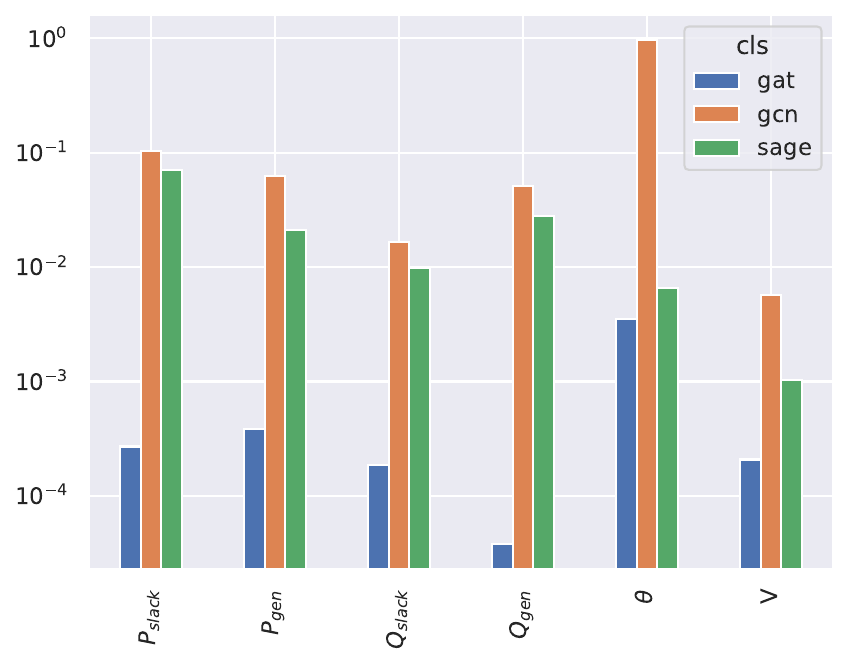}
    \caption{118-bus with ID variations}
    \label{fig:arch-case118-load}
\end{subfigure}
\begin{subfigure}{0.3\linewidth}
    \includegraphics[clip,width=\textwidth]{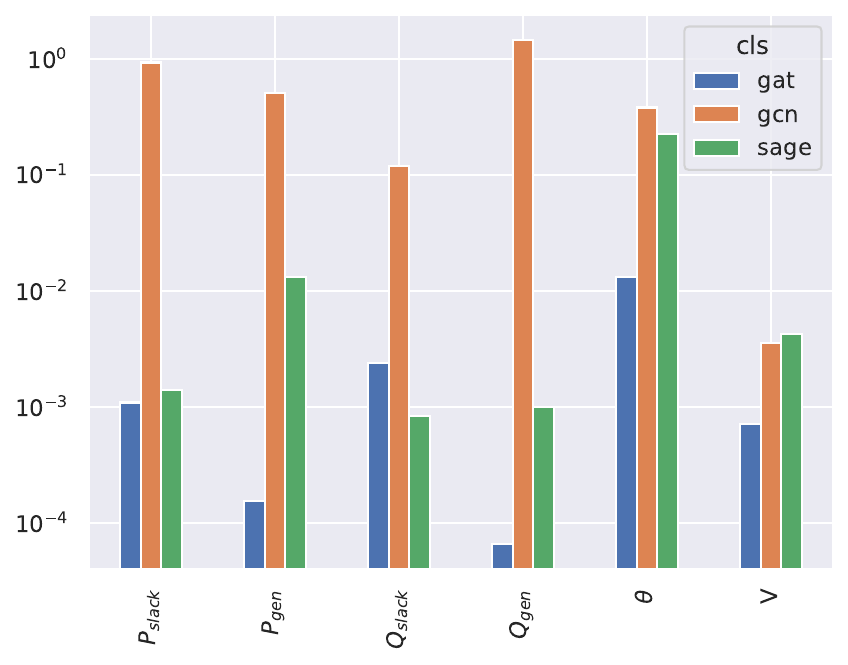}
    \caption{118-bus with line outages}
    \label{fig:arch-case118-line}
\end{subfigure}
\begin{subfigure}{0.3\linewidth}
    \includegraphics[clip,width=\textwidth]{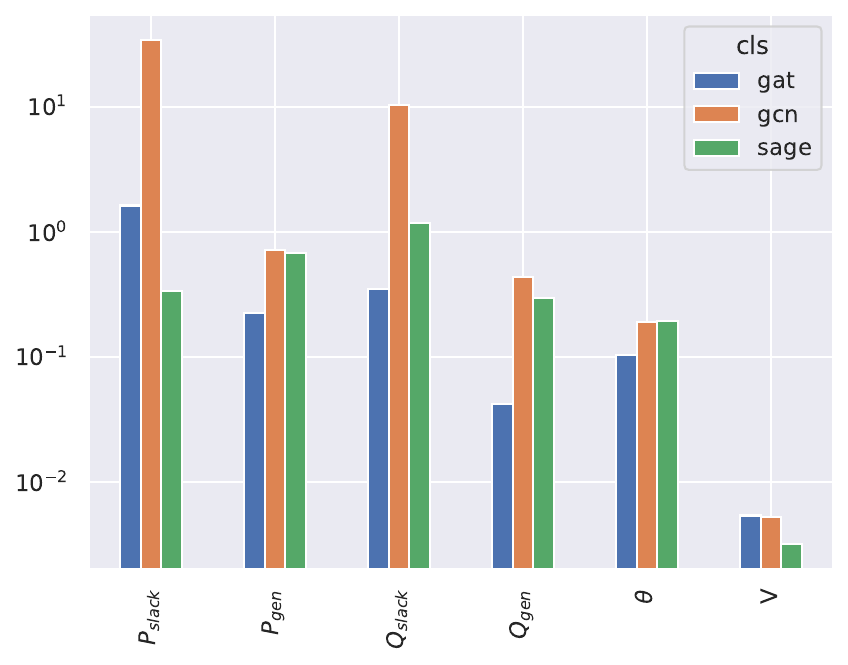}
    \caption{118-bus with price variations}
    \label{fig:arch-case118-cost}
\end{subfigure}
\caption{Impact of Architecture on the safety of the predictions.}
\label{fig:arch}
\end{figure*}

\textbf{Impact of GNN architecture.} We compare in Fig. \ref{fig:arch} the performance of three architectures, GCN, SAGE, and GAT over the ID cases, and their robustness to line outages and price variations. Across all sizes, scenarios and metrics, GAT is the best performing architecture. Across all scenarios, SAGE layers lead to the highest error in 15/18 metrics and 13/18 metrics for the 30-bus grids and the 9-bus grids respectively. For the large 118-bus grid, GCN is however the worst performing architecture for 16/18 metrics.

\begin{Insight}[boxed title style={colback=pink}]
GAT architecture is by far the best performing architecture for all scenarios across all grid sizes. The best performing architectures on ID scenarios remain the most robust (lowest errors) on perturbed scenarios.
\end{Insight}

\textbf{Impact of self-supervised learning.} We compare in Table \ref{tab:ssl} the performance of the best OPF GNN models with supervised learning alone and with a combination of supervised and self-supervised learning. We evaluate the ID scenario and the robustness to the price variations and line outages scenarios.

Our results show that combining SSL and SL improves the robustness to price variations on 4/6 metrics for the 9-bus grid. In other cases, SSL + SL achieves similar performance as SL alone, except for the Slack bus, where SL alone marginally outperforms SL+SSL.  


\begin{Insight}[boxed title style={colback=pink}]
Combining self-supervised learning with supervised learning for OPF learning can yield models robust against grid perturbations and as effective on ID variations as models trained solely with supervised learning.
\end{Insight}

\begin{table*}[ht]
\centering
\caption{Impact of Self-supervised learning on the robustness of the OPF predictions: Normalized SE value for each component in the form: mean (std). Lower values are better. In bold the best cases.}
\label{tab:ssl}
\resizebox{\textwidth}{!}{
\begin{tabular}{c|c|c|cccccc}
\toprule
Case & Var    &  Learning    & P\_gen                                                  & Q\_gen                                                   & P\_slack            & Q\_slack                                                & V                                                       & $\theta$                                                     \\ 
\midrule

9-bus   & ID           & SL       & \num{0.000020} (\num{0.000021})         & \num{0.0000015} (\num{0.0000021})       & \num{9.961866e-06}(\num{0.000009})          & \num{0.000006}(\num{0.000009})                    & \num{0.000046}(\num{0.000055})   & \num{0.000137}(\num{0.000233})           \\
        &              & +SSL     & \num{0.0000010} (\num{0.000013})        & \num{0.0000011} (\num{0.000014})        & \textbf{\num{8.561689e-07}(\num{0.000001})} & \num{0.000005}(\num{0.000005})                    & \num{0.000044}(\num{0.000056})   & \num{0.000036}(\num{0.000043})           \\ \cline{2-9} 
        & Price        & SL       & \num{0.174139}(\num{0.279312})          & \num{0.103441}(\num{0.260729})          & \num{0.121786}(\num{0.115199})              & \num{0.031686}(\num{0.054420})                    & \num{0.002471}(\num{0.005689})   & \num{0.023571}(\num{0.038714})           \\
        &              & +SSL     & \textbf{\num{0.094627}(\num{0.079535})} & \textbf{\num{0.003559}(\num{0.025835})} & \textbf{\num{0.048768}(\num{0.047070})}     & \textbf{\num{0.012945}(\num{0.014138})}           & \num{0.003550}(\num{0.007067})   & \num{0.023814}(\num{0.045827})           \\ \cline{2-9} 
        & Line         & SL       & \num{0.000522}(\num{0.001080})          & \textbf{\num{0.001006}(\num{0.001665})} & \textbf{\num{0.000032}(\num{0.000037})}     & \num{0.001992}(\num{0.005367})                    & \num{0.000641}(\num{0.001248})   & \num{0.008957}(\num{0.016149})           \\
        &              & +SSL     & \num{0.000273}(\num{0.000380})          & \num{0.003120}(\num{0.009213})          & \num{0.000083}(\num{0.000123})              & \num{0.003093}(\num{0.006462})                    & \num{0.000718}(\num{0.001527})   & \num{0.008625}(\num{0.015668})           \\

      \midrule
      
30-bus  & ID           & SL       & \num{0.000004} (\num{0.000010})         & \num{0.000006} (\num{0.000011})         & \num{9.243440e-07}(\num{0.000001})          & \num{1.004663e-06} (\num{2.453581e-06})           & \num{0.000061} (\num{0.000109}) & \num{0.000085} (\num{0.000500})          \\
        &              & +SSL     & \num{0.000003} (\num{0.000009}         & \num{0.000002} (\num{0.000006}         & \num{1.001064e-06}(\num{0.000004})          & \textbf{\num{1.488563e-07} (\num{3.556052e-07})} & \num{0.000064}(\num{0.000116})   & \num{0.000115} (\num{0.000607})          \\ \cline{2-9} 
        & Price - sage & SL       & \num{0.004603}(\num{0.007646})          & \num{0.000841}(\num{0.002431})          & \textbf{\num{0.000449}(\num{0.000509})}     & \num{0.000294}(\num{0.001292})                    & \num{0.000184}(\num{0.000418})   & \num{0.000666}(\num{0.001562})           \\
        &              & +SSL     & \num{0.011085}(\num{0.028624})          & \num{0.001320}(\num{0.004620})          & \num{0.003921}(\num{0.007557})              & \num{0.000364}(\num{0.001289})                    & \num{0.000163}(\num{0.000425})   & \num{0.000771}(\num{0.001680})           \\ \cline{2-9} 
        & Line         & SL       & \num{0.000029}(\num{0.000349})          & \num{0.000045} (\num{0.000968})         & \num{0.000004}(\num{0.000001})              & \num{1.461339e-06}(\num{3.053494e-06})            & \num{0.000132}(\num{0.000474})   & \num{0.000156}(\num{0.000849})           \\
        &              & +SSL     & \num{0.000024}(\num{0.000197})          & \num{0.000018}(\num{0.000047})          & \num{0.000002}(\num{0.000007})              & \textbf{\num{5.813202e-07}(\num{8.033968e-07})}   & \num{0.000126}(\num{0.000480})   & \num{0.000174}(\num{0.000986})           \\

\midrule
118-bus & ID           & SL       & \num{0.000387} (\num{0.001152})         & \num{0.000028} (\num{0.000079})         & \textbf{\num{0.000040}(\num{0.000021})}     & \num{0.000187} (\num{0.000115})                   & \num{0.000208} (\num{0.000441})  & \textbf{\num{0.003209} (\num{0.006585})} \\
        &              & +SSL     & \num{0.000361} (\num{0.000908}         & \num{0.000027} (\num{0.000091})         & \num{0.000140}(\num{0.000085})              & \textbf{\num{0.000021} (\num{0.000029})}          & \num{0.000648}(\num{0.001379})   & \num{0.020197} (\num{0.112317})          \\ \cline{2-9} 
        & Price        & SL       & \num{0.212251}(\num{0.373310})          & \num{0.036437}(\num{0.118089})          & \textbf{\num{0.336550}(\num{0.291254})}     & \num{0.548407}(\num{0.601089})                    & \num{0.003203}(\num{0.004565})   & \num{0.191129}(\num{0.452715})           \\
        &         & +SSL     & \num{0.263795}(\num{0.474748})          & \num{0.057357}(\num{0.224172})          & \num{2.895216} (\num{3.588339})              & \num{0.853425}(\num{0.951980})                    & \num{0.003571}(\num{0.004533})   & \num{0.334123}(\num{0.564933})           \\ \cline{2-9} 
        & Line         & SL       & \num{0.000157} (\num{0.001153})         & \num{0.000066} (\num{0.000620})         & \num{0.000037} (\num{0.000142})             & \num{0.000655} (\num{0.001639})                   & \num{0.000719} (\num{0.001994})  & \num{0.001956} (\num{0.005558})          \\
        &              & +SSL     & \num{0.000551} (\num{0.005674})         & \num{0.000226} (\num{0.004566})         & \textbf{\num{0.000015} (\num{0.000019})}    & \num{0.000144} (\num{0.000618})                   & \num{0.000250} (\num{0.000700})  & \num{0.001851} (\num{0.011665})          \\

 \bottomrule                                 
\end{tabular}
}
\end{table*}

\subsection{OPF qualitative evaluation}

\textbf{How stable is our framework across simulators?} 

We evaluate the robustness of the oracle by comparing the OPF solution from the three simulators Pandapower, MatPower and PowerModels.jl. We mutate the initial grid to generate 100 mutants for each grid size. 

First, we compare the solutions found by each simulator when changing the initial states of the grid with 10 random seeds, and the standard deviation of the solvers solutions remained zero.

\begin{figure}[t]
\centering
\begin{subfigure}{0.45\linewidth}
    \includegraphics[clip, trim={0px 0px 250px 0px}, width=\textwidth]{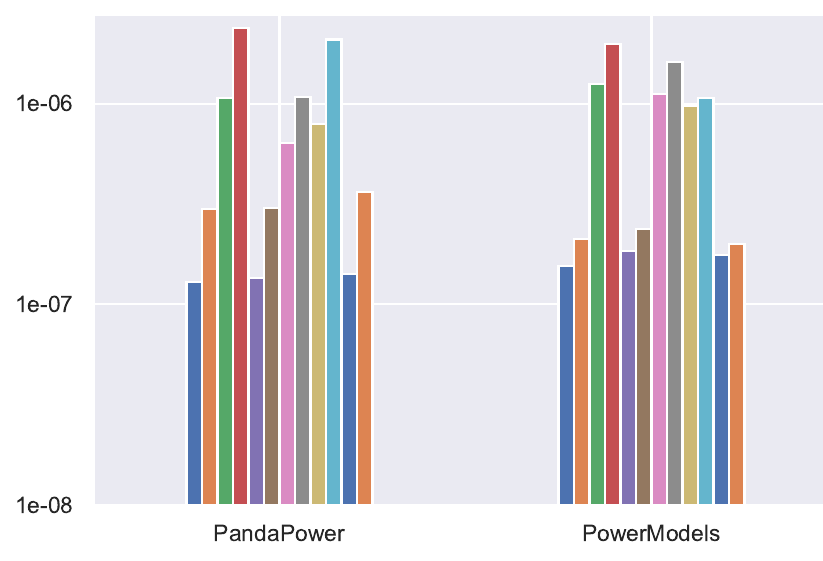}
    \caption{9-bus - ID}
    \label{fig:simulators-case9-id}
\end{subfigure}
\begin{subfigure}{0.45\linewidth}
    \includegraphics[clip, trim={0px 0px 250px 0px}, width=\textwidth]{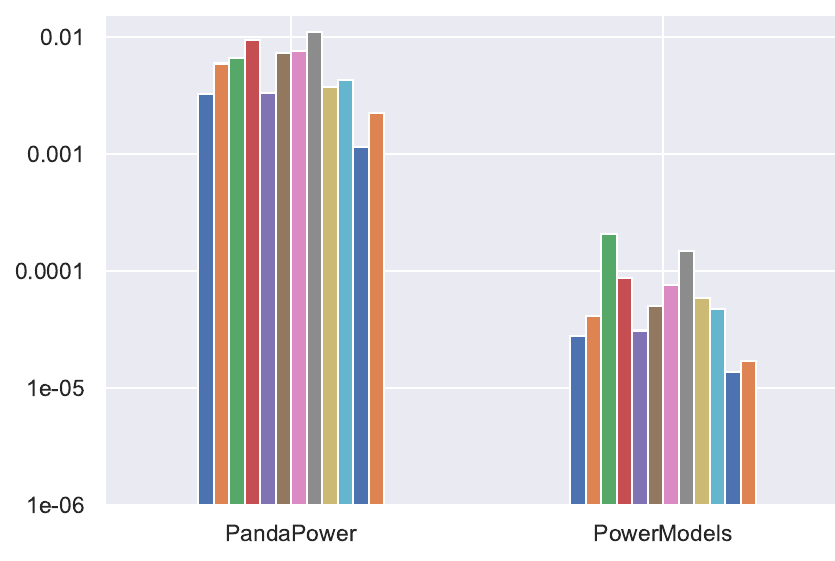}
    \caption{30-bus - ID}
    \label{fig:simulators-case30-id}
\end{subfigure}
\begin{subfigure}{0.45\linewidth}
    \includegraphics[clip, trim={0px 0px 250px 0px}, width=\textwidth]{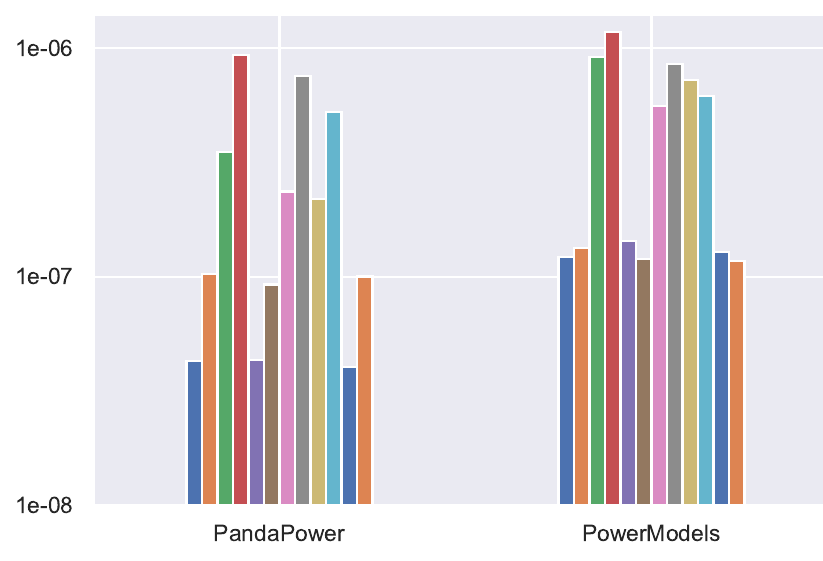}
    \caption{price variations}
    \label{fig:simulators-case9-cost}
\end{subfigure}
\begin{subfigure}{0.45\linewidth}
    \includegraphics[clip, trim={0px 0px 250px 0px}, width=\textwidth]{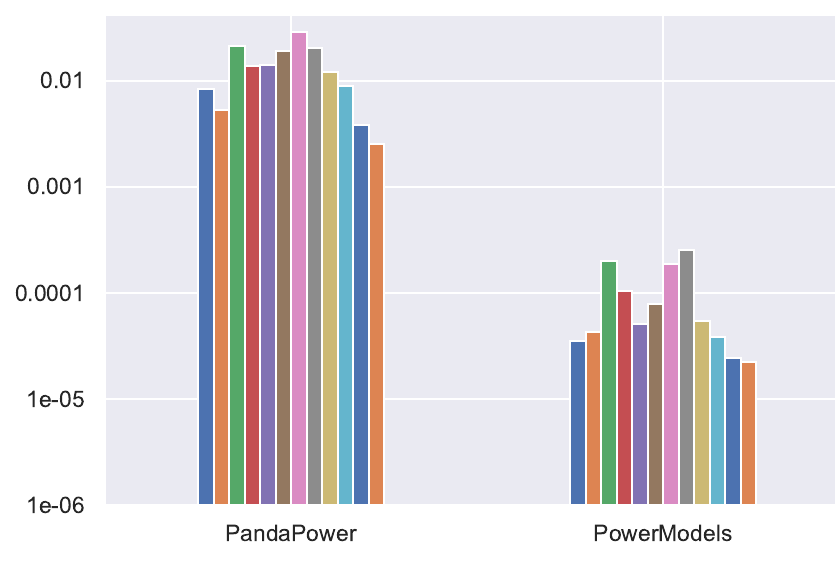}
    \caption{30-bus - Price variations}
    \label{fig:simulators-case30-cost}
\end{subfigure}
\begin{subfigure}{0.45\linewidth}
    \includegraphics[clip, trim={0px 0px 250px 0px}, width=\textwidth]{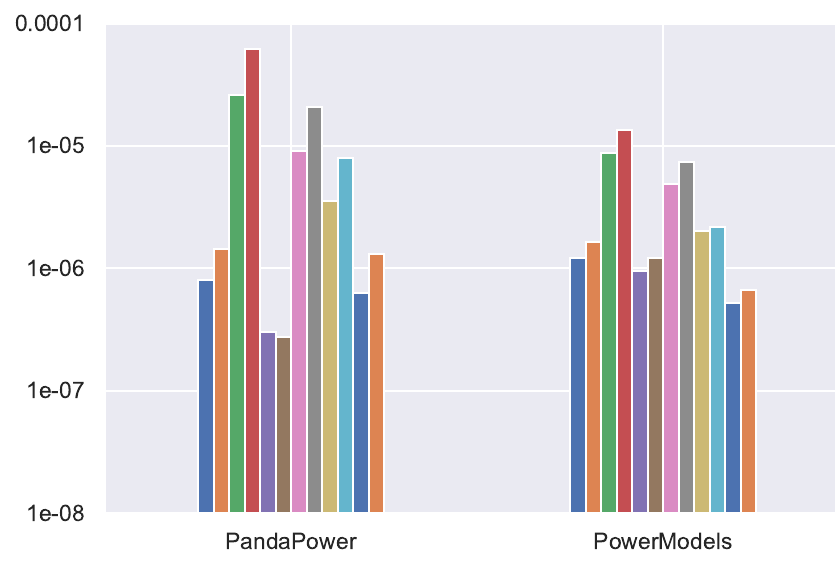}
    \caption{9-bus - Line outages}
    \label{fig:simulators-case9-line}
\end{subfigure}
\begin{subfigure}{0.45\linewidth}
    \includegraphics[clip, trim={0px 0px 250px 0px}, width=\textwidth]{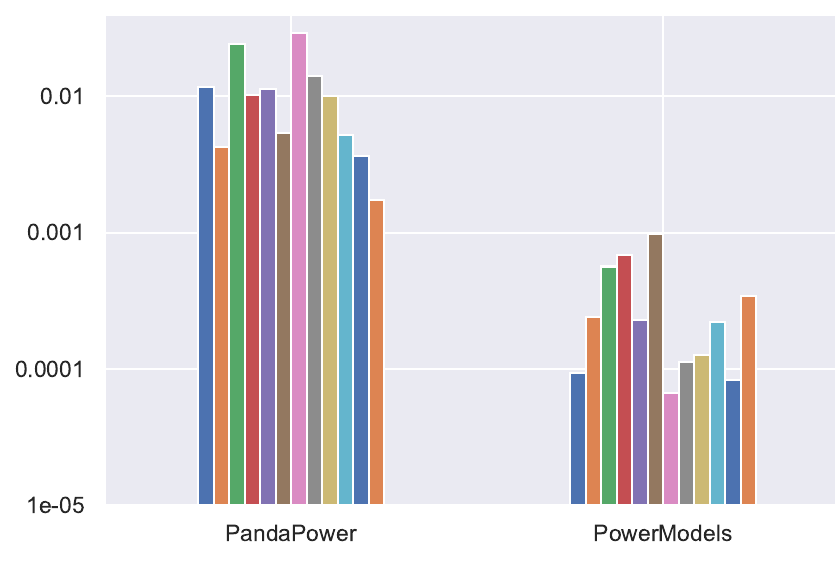}
    \caption{30-bus - Line outages}
    \label{fig:simulators-case30-line}
\end{subfigure}
\begin{subfigure}{0.45\linewidth}
    \includegraphics[clip, trim={0px 0px 250px 0px},width=\textwidth]{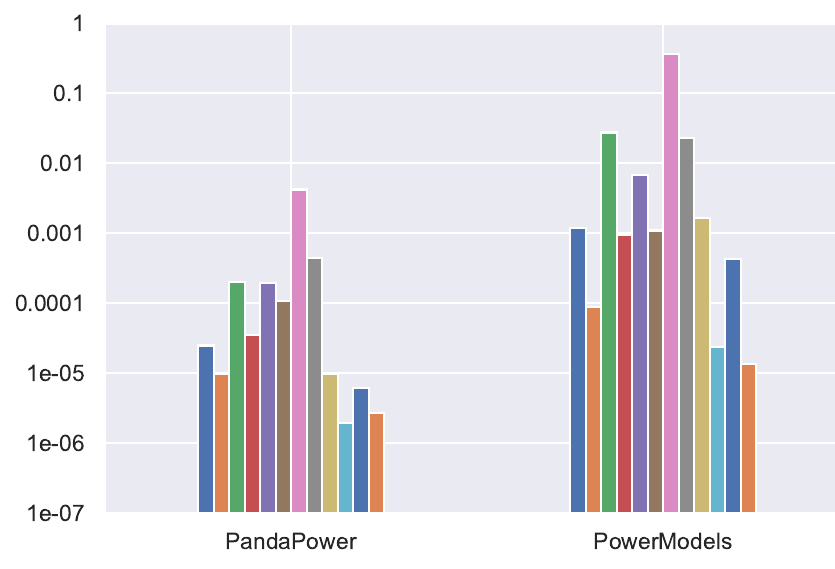}
    \caption{118-bus - ID}
    \label{fig:simulators-case118-id}
\end{subfigure}
\begin{subfigure}{0.45\linewidth}
    \includegraphics[clip, trim={0px 0px 250px 0px},width=\textwidth]{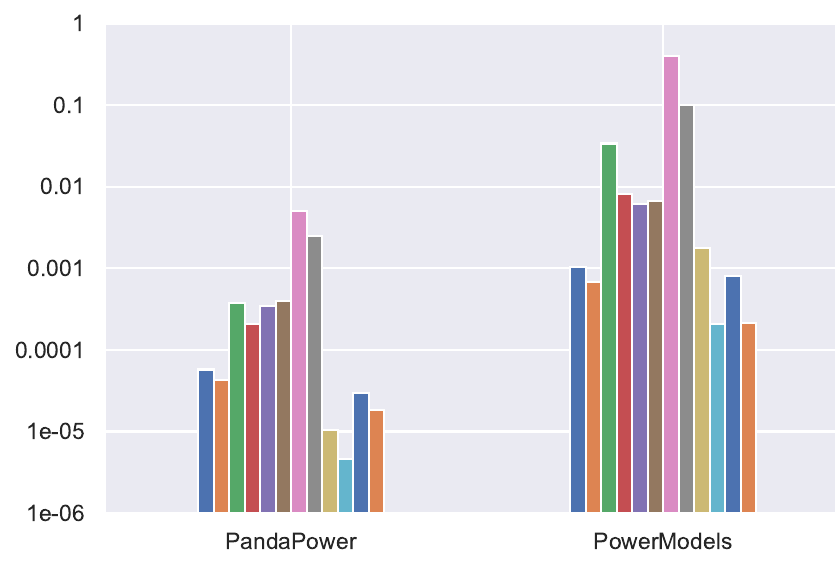}
    \caption{118-bus - Price variations}
    \label{fig:simulators-case118-cost}
\end{subfigure}
\begin{subfigure}{0.45\linewidth}
    \includegraphics[clip, trim={0px 0px 250px 0px},width=\textwidth]{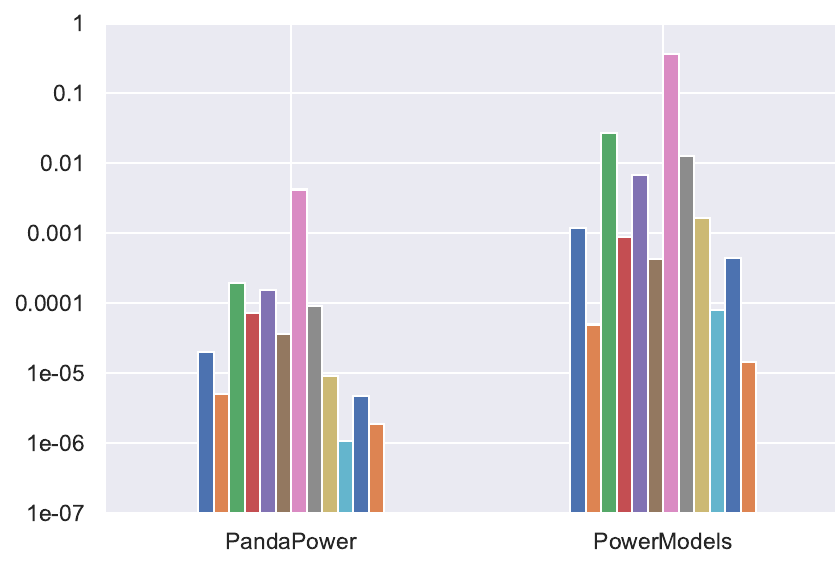}
    \caption{118-bus - Line outages}
    \label{fig:simulators-case118-line}
\end{subfigure}
\begin{subfigure}{0.45\linewidth}
    \includegraphics[clip, trim={405px 20px 0px 60px},width=\textwidth]{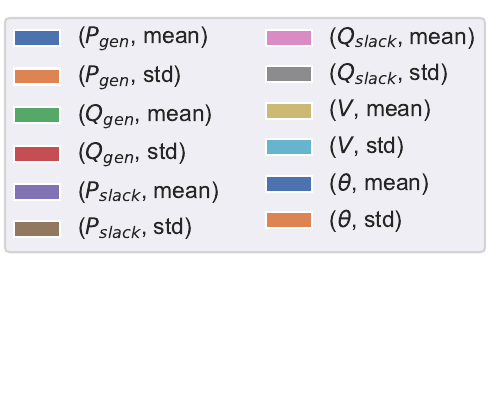}
\end{subfigure}
\hfill

\caption{Impact of the simulators on the robustness of the oracle for OPF. Relative MSE between the solutions by MatPower and both Pandapower and PowerModels.jl.}
\label{fig:simulators}
\end{figure}

Next, we compare the solutions found by each simulator, and we report the mean and standard deviation of the difference to MatPower solutions, of the solutions from Pandapower and PowerModels.jl respectively in Fig. \ref{fig:simulators}.

Our results show that (1) our framework yields similar ground-truth labels for the three OPF simulation tools on the ID scenario with $\leq \num{2e-6}$ and $\leq \num{5e-2}$ errors across all metrics and simulators on the 9-bus grids and 30-bus grids, respectively. 

(2) Across all grid sizes, there is no significant variation in simulators' errors between the ID scenario and the perturbed scenarios except for the 9-bus grid with line outages where the errors increase a up to hundred-fold compared to ID. On this small grid, line outages have a major impact, and simple simulators such as pandapower may underform compared to Matpower. 
In all cases, the errors induced by grid perturbation on the GNN predictions (presented in Table \ref{tab:ssl}). 

\begin{Insight}[boxed title style={colback=pink}]
SafePowerGraph is reliable and simulator-agnostic: Its GNN solutions yield lower errors or of similar magnitude as the errors between different solvers' solutions. 
\end{Insight}

\textbf{Which buses and constraints are most sensitive?} 

\begin{figure*}[t]
\centering
\begin{subfigure}{0.3\linewidth}
    \includegraphics[clip, width=\textwidth]{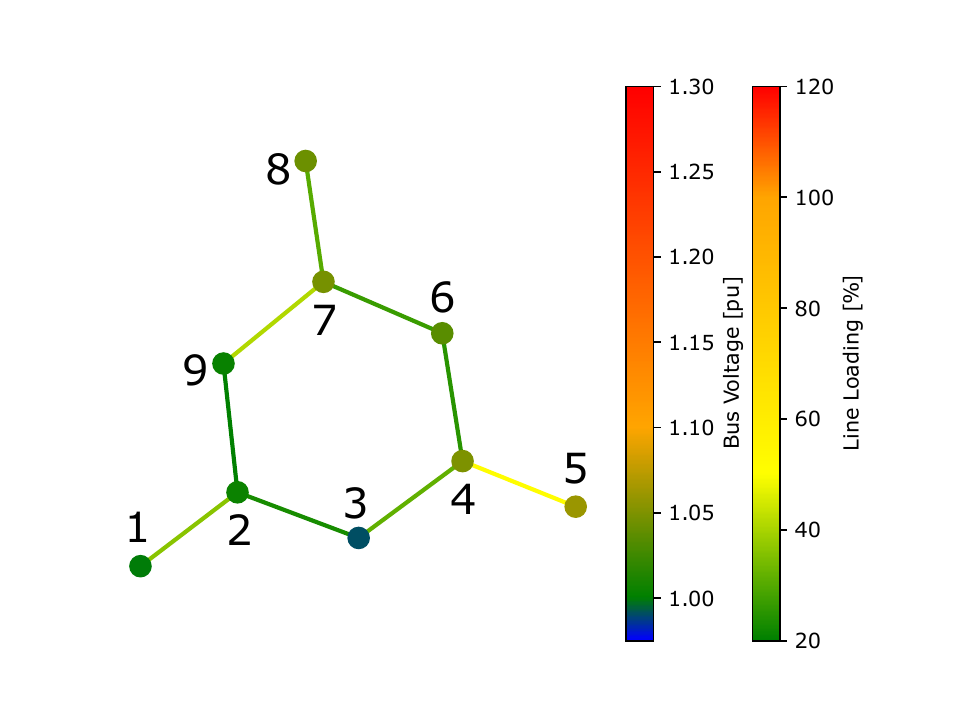}
    \caption{Initial 9-bus}
    \label{fig:line-case9-grid}
\end{subfigure}
\begin{subfigure}{0.22\linewidth}
    \includegraphics[clip, trim={0px 0px 180px 0px}, width=\textwidth]{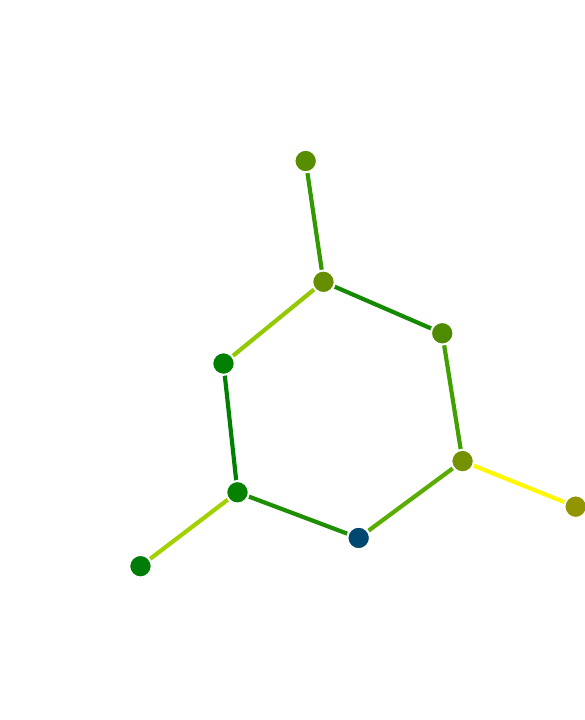}
    \caption{price variation: Oracle}
    \label{fig:line-case9-oracle5}
\end{subfigure}
\begin{subfigure}{0.22\linewidth}
    \includegraphics[clip, trim={0px 0px 180px 0px}, width=\textwidth]{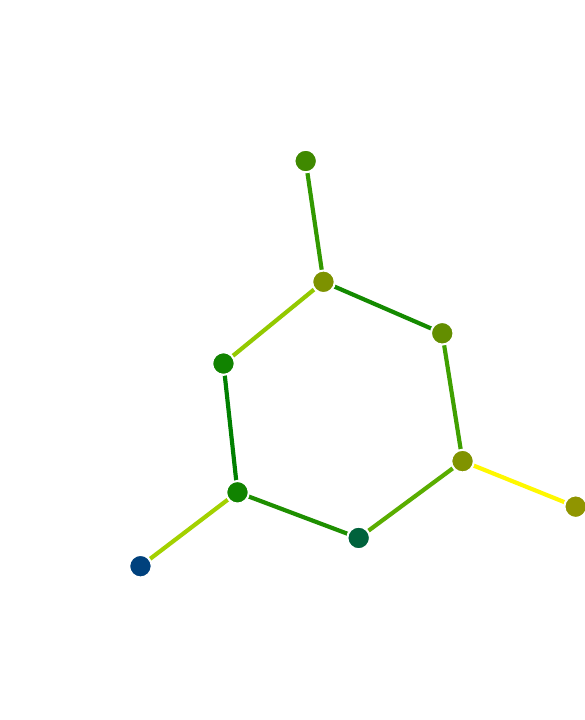}
    \caption{price variation: GNN}
    \label{fig:line-case9-gnn5}
\end{subfigure}
\hfill
\begin{subfigure}{0.3\linewidth}
     \centering
        \rule{\textwidth}{0pt}
\end{subfigure}
\begin{subfigure}{0.22\linewidth}
    \includegraphics[clip, trim={0px 0px 180px 0px}, width=\textwidth]{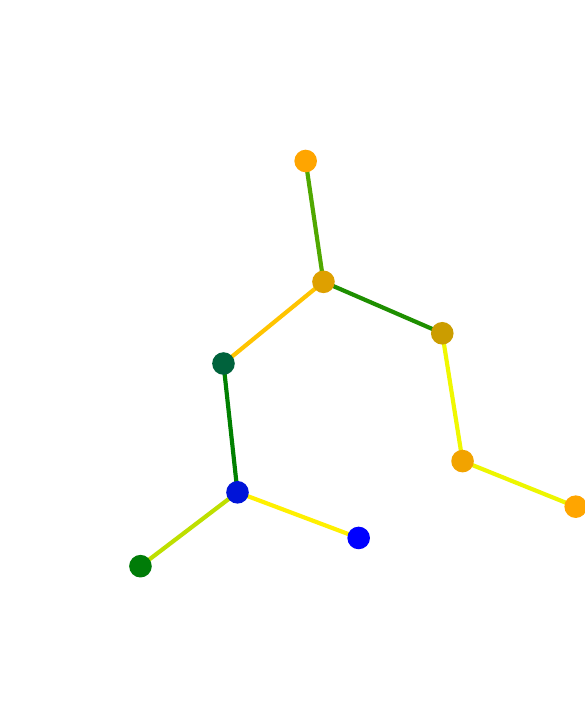}
    \caption{line 7 outage: Oracle}
    \label{fig:line-case9-oracle7}
\end{subfigure}
\begin{subfigure}{0.22\linewidth}
    \includegraphics[clip, trim={0px 0px 180px 0px}, width=\textwidth]{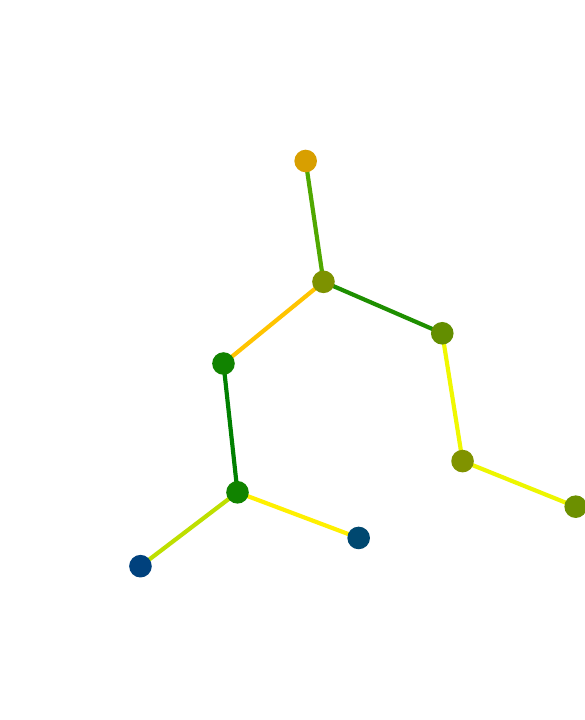}
    \caption{line 7 outage: GNN}
    \label{fig:line-case9-gnn7}
\end{subfigure}
\hfill
\begin{subfigure}{0.3\linewidth}
    \includegraphics[clip, width=\textwidth]{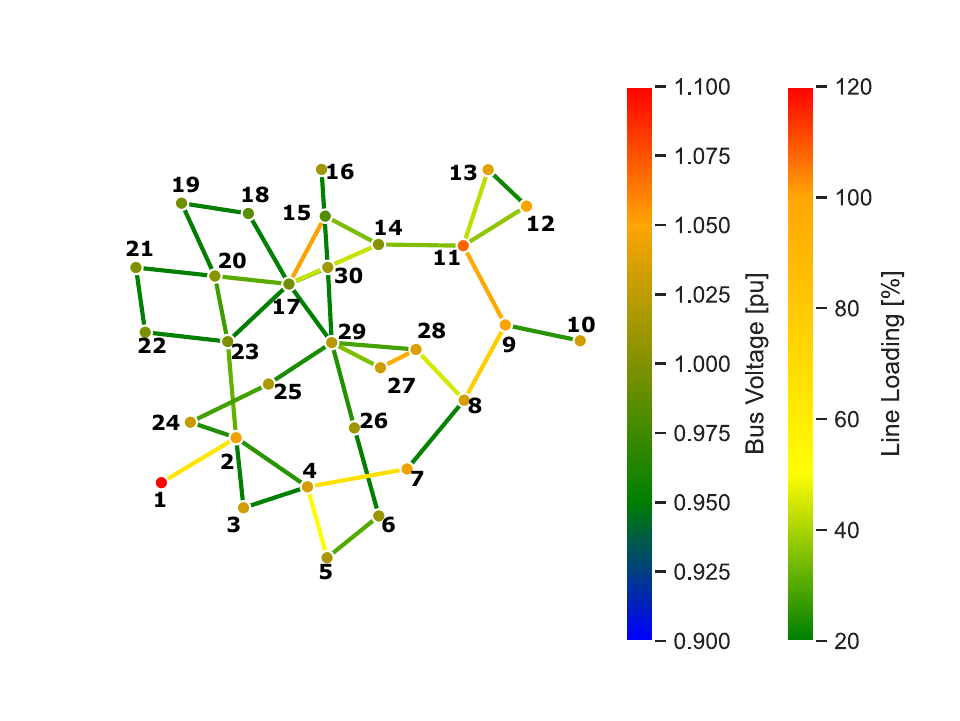}
    \caption{30-bus grid before line outage}
    \label{fig:line-case30}
\end{subfigure}
\begin{subfigure}{0.22\linewidth}
    \includegraphics[clip, trim={0px 0px 180px 0px}, width=\textwidth]{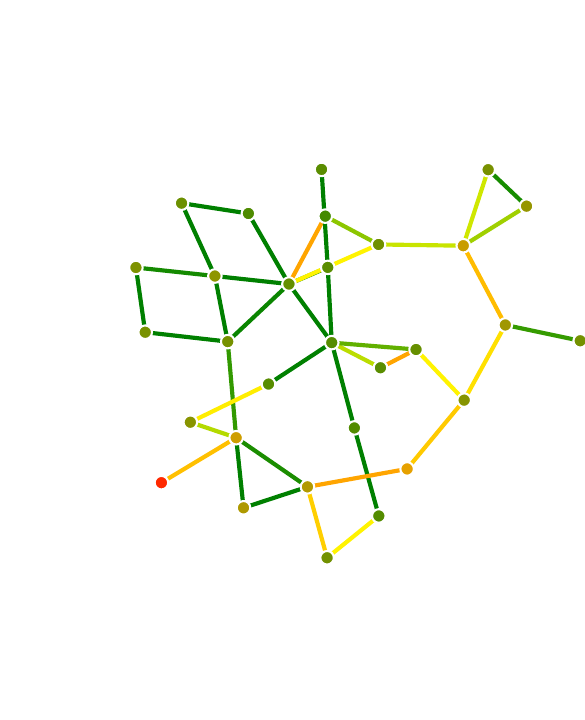}
    \caption{price variation: Oracle}
    \label{fig:line-case30-oraclePrice}
\end{subfigure}
\begin{subfigure}{0.22\linewidth}
    \includegraphics[clip, trim={0px 0px 180px 0px}, width=\textwidth]{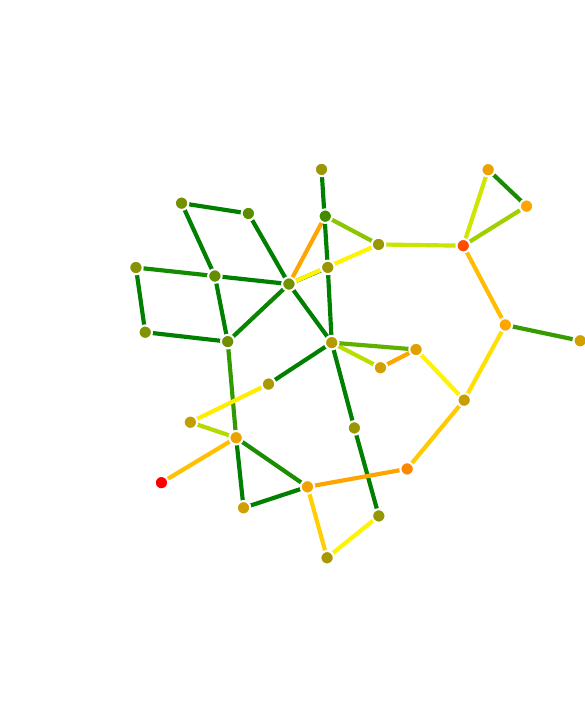}
    \caption{price variation: GNN}
    \label{fig:line-case30-gnnPrice}
\end{subfigure}\hfill
\begin{subfigure}{0.3\linewidth}
    \rule{\textwidth}{0pt}
\end{subfigure}
\begin{subfigure}{0.22\linewidth}
    \includegraphics[clip, trim={0px 0px 180px 0px}, width=\textwidth]{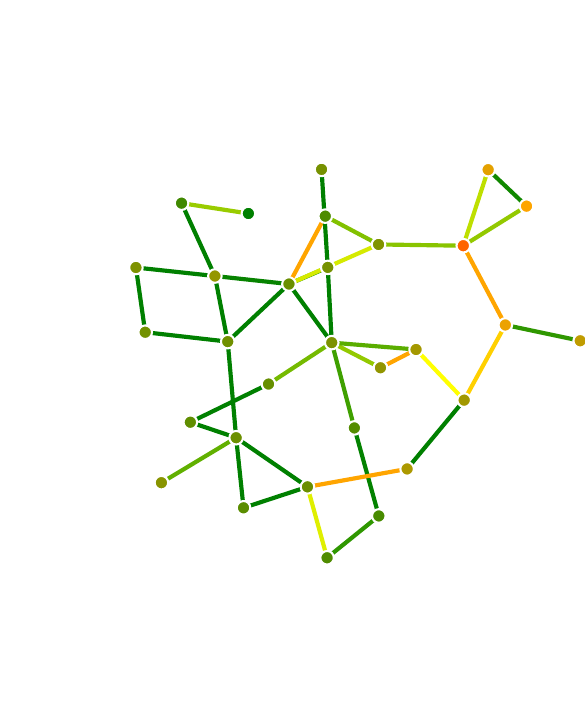}
    \caption{line 8 outage: Oracle}
    \label{fig:line-case30-oracle8}
\end{subfigure}
\begin{subfigure}{0.22\linewidth}
    \includegraphics[clip, trim={0px 0px 180px 0px}, width=\textwidth]{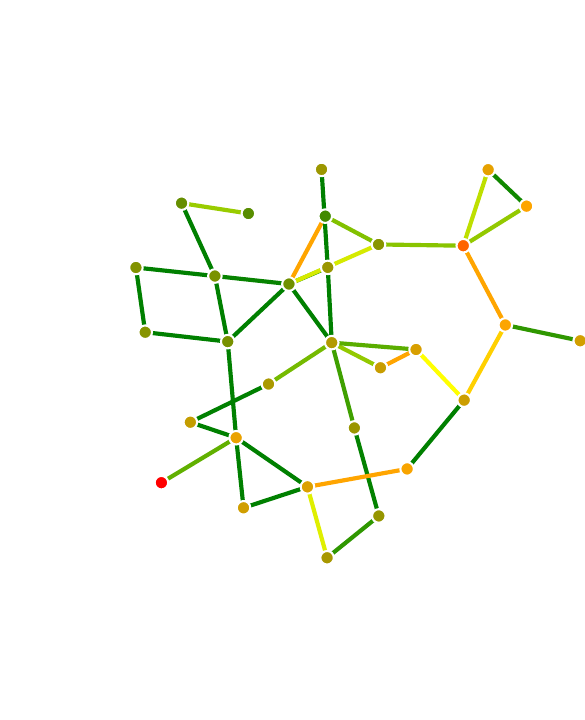}
    \caption{line 8 outage: GNN}
    \label{fig:line-case30-gnn8}
\end{subfigure}

\caption{Impact of line outage and price perturbation on the safety of the OPF predictions.}
\label{fig:lines}
\end{figure*}

We present in Fig. \ref{fig:lines} examples of 9-bus and 30-bus grids under perturbations. We report the Bus voltage in p.u (i.e. the loading of the buses). Ideal values should be between 0.95 and 1.05, and unsafe values are above 1.10. 
For the 9-bus, the solutions predicted by GNN closely match the oracle at individual bus levels for price variations. There are however more differences between the buses 1,2, and 3  between the oracle and the GNN predictions when the line 7 is cut (Fig. \ref{fig:line-case9-oracle7} and Fig. \ref{fig:line-case9-gnn7}).  

On the 30-bus grid, there is limited impact to line outages; for example, line 8 outage in Fig. \ref{fig:line-case30-oracle8}. Under price variations, the nodes connected to bus 28 and bus 21 show the largest differences between the oracle and the GNN predictions.

\begin{Insight}[boxed title style={colback=pink}]
Overall, at the individual bus levels, the solutions found by the GNN trained with SafePowerGraph vary from the oracle solutions. While being outside the ideal bounds they remain however within the safety range. 
\end{Insight}

\begin{figure}[t]
\centering
\begin{subfigure}{\linewidth}
    \includegraphics[clip, width=\textwidth]{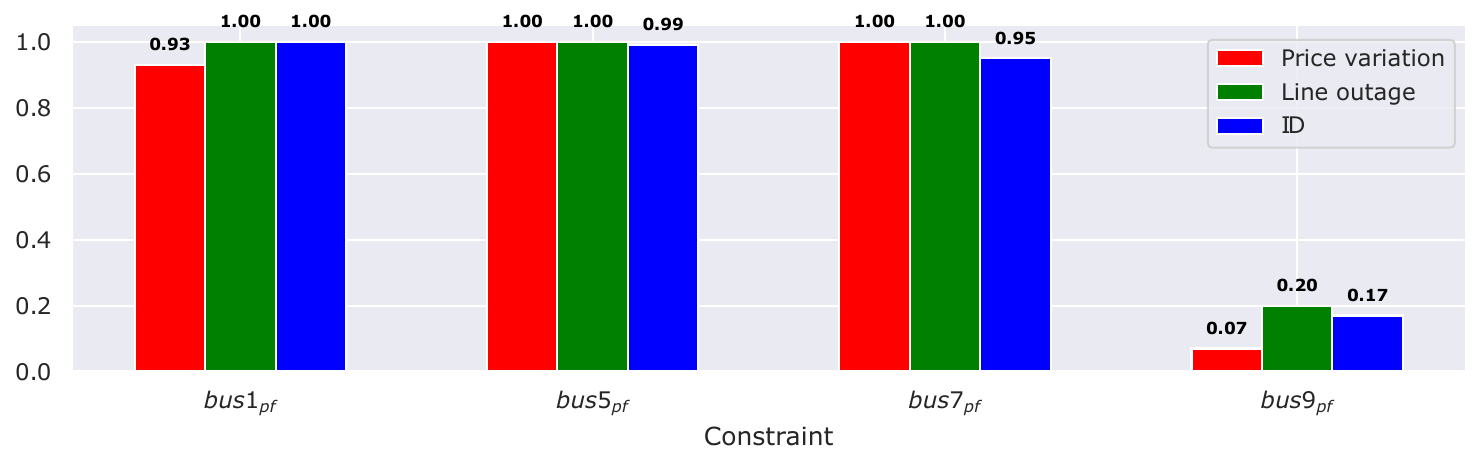}
    \caption{9-bus}
    \label{fig:ctrs-case9}
\end{subfigure}
\hfill
\begin{subfigure}{\linewidth}
    \includegraphics[clip, width=\textwidth]{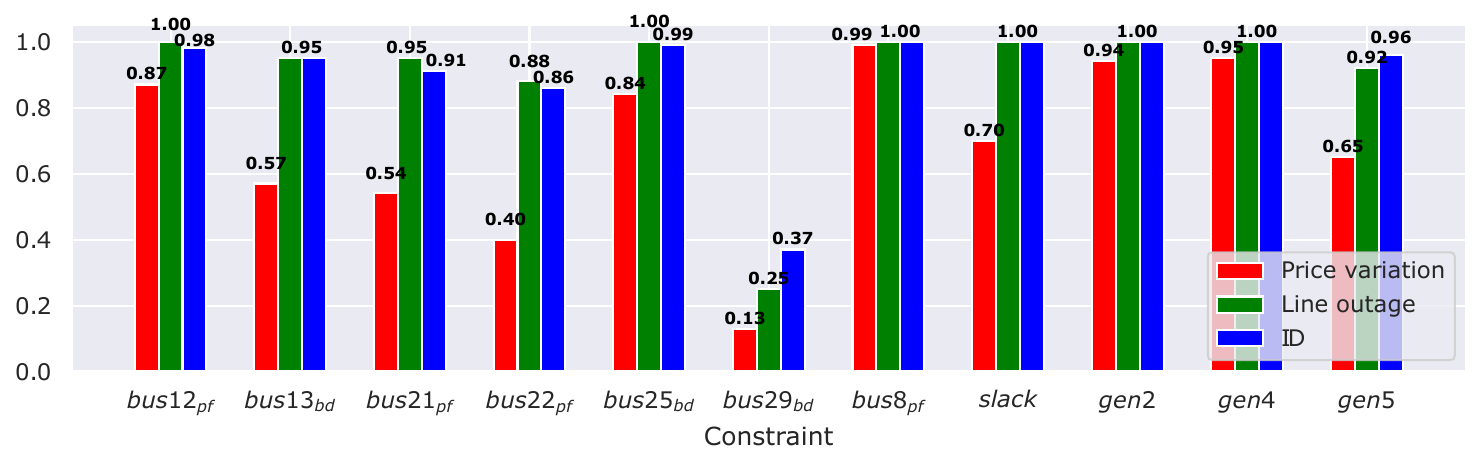}
    \caption{30-bus}
    \label{fig:ctrs-case30}
\end{subfigure}

\caption{Impact of line outages and price variations on the constraints satisfaction. "$busX_{pf}$" refer to power flow constraints off bus X, and to "$busX_{bd}$"  its boundary constraints. There are 21 and 66 constraints in total for the 9-bus and the 30-bus grids respectively. Only broken constraints are displayed. The numbering of the nodes starts at 1.}
\label{fig:ctrs}
\end{figure}

Next, we investigate the individual nodes and  evaluate in Fig. \ref{fig:ctrs} for each component of the grids, which constraints are satisifed under different scenarios. 
The 9-bus grid has 21 constraints while the 30-bus grid has 66 constraints.

For the 9-bus grid, except for the powerflow constraint of bus 9, all the constraints are satisfied on ID and line outage cases. On the 30-bus grid, buses 12, 13, 21, 22, 25, and 29, and generators 2,4, and 5 do not satisfy their constraints on some of the validation graphs. 
Under price variations, the powerflow constraints of bus 21 are not satisfied in more than 60\% of the predicted graph solutions, and the boundary constraints of bus 28 are not satisifed in more than 85\% of the grids. 

Considering violations occurring in more than 5\% of the predicted grids: under price variations, two constraints are broken for 9-bus, nine for 30 bus grids. Meanwhile, predictions with line outages cause one constraint violation on 9-bus, and two for 30-bus grids. 

\begin{Insight}[boxed title style={colback=pink}]
Significantly more powerflow and boundary constraints are broken by the predicted solutions under price variations scenarios than under line outages scenarios. 
\end{Insight}

\subsection{Powerflow evaluation}

\begin{figure}[H]
\centering
\begin{subfigure}{.80\linewidth}
    \includegraphics[clip, width=\textwidth]{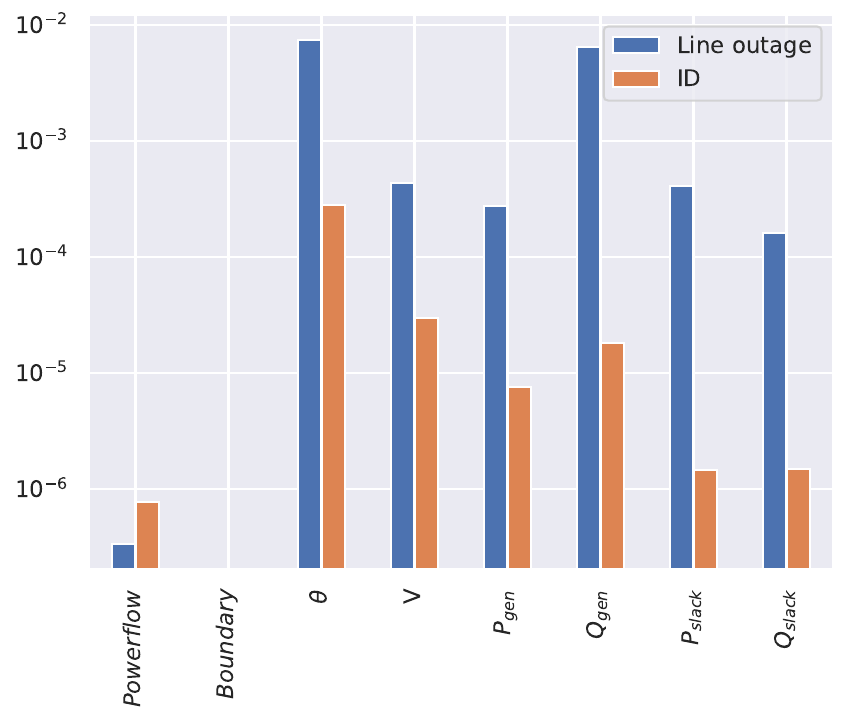}
    \caption{9-bus}
    \label{fig:variations-pf-case9}
\end{subfigure}
\begin{subfigure}{.80\linewidth}
    \includegraphics[clip, width=\textwidth]{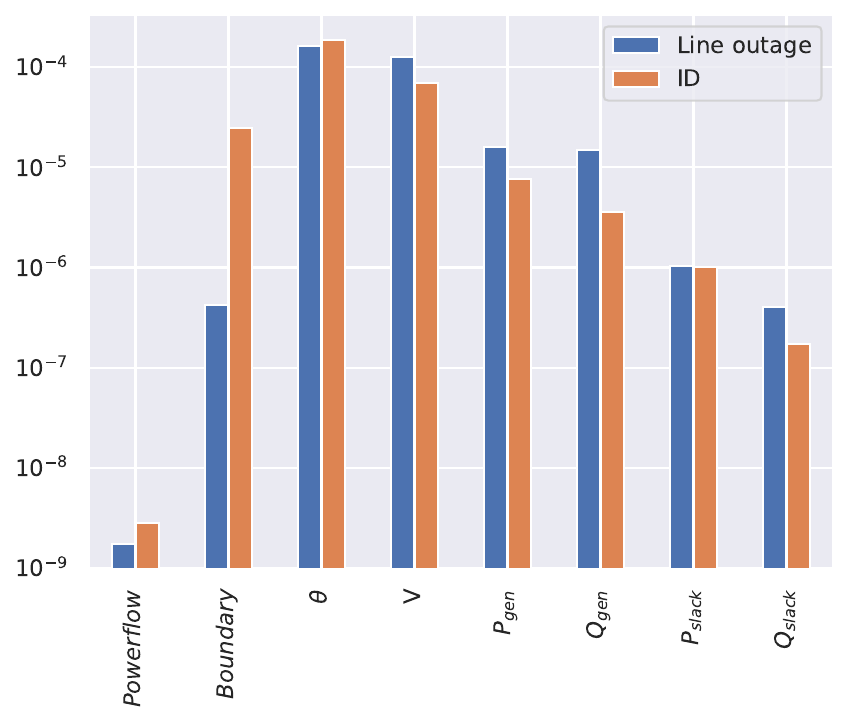}
    \caption{30-bus}
    \label{fig:variations-pf-case30}
\end{subfigure}

\caption{Impact of power grid perturbations on the robustness and safety of the OPF predictions.}
\label{fig:variations-pf}
\vspace{-1em}
\end{figure}

The PF problem only supports line outage perturbations. In the following we report the results for the 9-bus and 30-bus cases, and provide the detailed results (including the 118-bus grid) in Appendix \ref{sec:app-B}.

\textbf{Impact of grid perturbations}

We evaluated the impact of grid perturbations on the supervised errors and the grid constraints errors in Fig. \ref{fig:variations-pf}. On the small 9-bus grid, the problem can be solved with exactly zero boundary violation. Although line outages lead to higher errors than the ID scenario on the 9-bus grid, both scenarios achieve similar errors on the 30-bus grid. 

\begin{figure}[H]
\centering
\begin{subfigure}{0.49\linewidth}
    \includegraphics[clip, width=\textwidth]{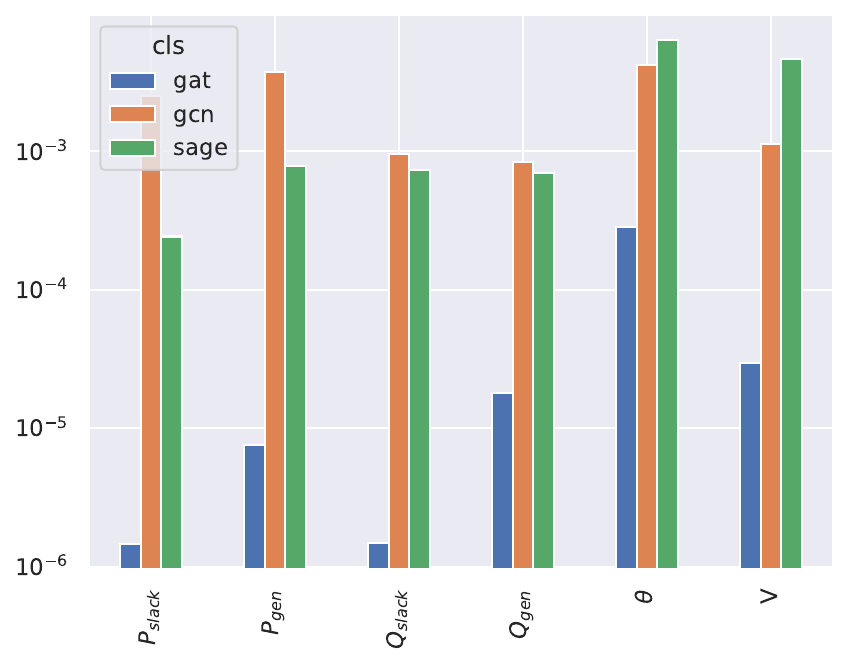}
    \caption{9-bus with ID variations}
    \label{fig:arch-pf-case9-load}
\end{subfigure}
\begin{subfigure}{0.49\linewidth}
    \includegraphics[clip, width=\textwidth]{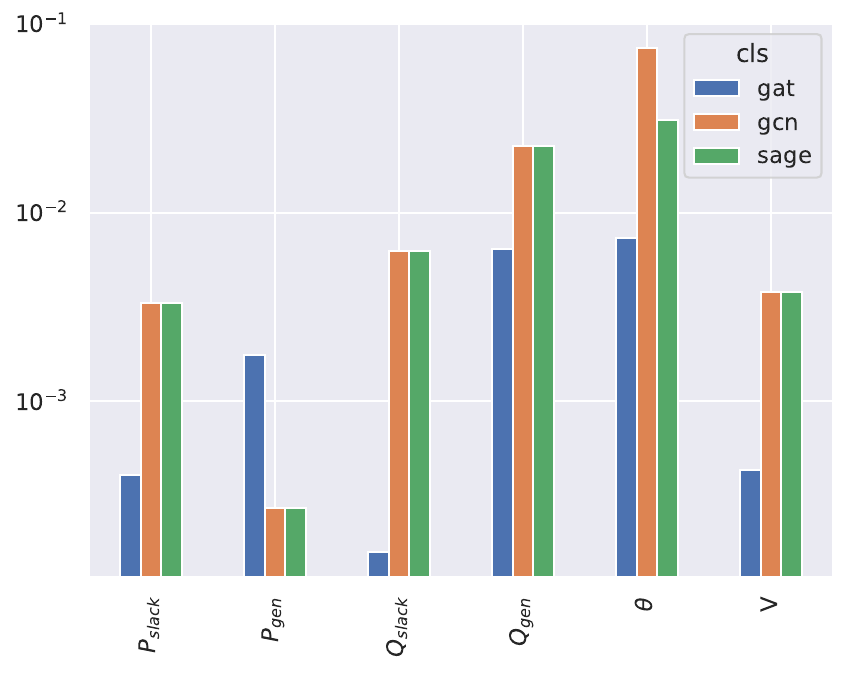}
    \caption{9-bus with line outages}
    \label{fig:arch-pf-case9-line}
\end{subfigure}
\hfill
\begin{subfigure}{0.49\linewidth}
    \includegraphics[clip, width=\textwidth]{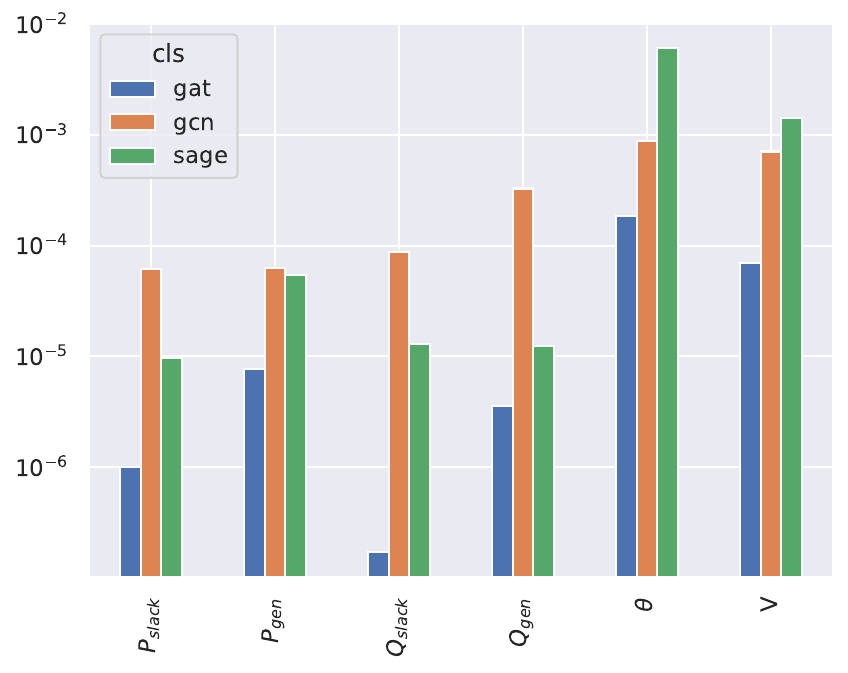}
    \caption{30-bus with ID variations}
    \label{fig:arch-pf-case30-load}
\end{subfigure}
\begin{subfigure}{0.49\linewidth}
    \includegraphics[clip,width=\textwidth]{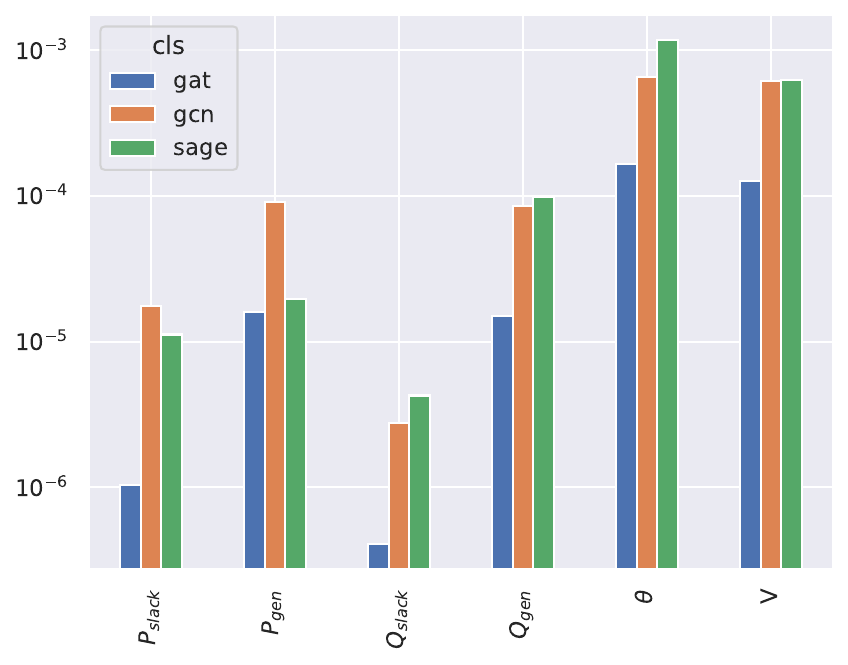}
    \caption{30-bus with line outages}
    \label{fig:arch-pf-case30-line}
\end{subfigure}

\caption{Impact of Architecture on the safety of the Powerflow predictions.}
\label{fig:arch-pf}
\end{figure}

\textbf{Impact of architecture}
We evaluate in Fig. \ref{fig:arch-pf} the impact of architectures on the performance and robustness of the models.
Our results confirm our previous insight that GAT is also the best architecture across grid perturbations for the PF problem.

\begin{table*}[t]
\centering
\vspace{1em}
\caption{Impact of Self-supervised learning on the robustness of the Powerflow predictions: Normalized SE value for each component in the form: mean (std). Lower values are better. In bold the best cases.}
\label{tab:pf-ssl}
 \resizebox{\textwidth}{!}{
\begin{tabular}{c|c|c|cccccc}
\toprule
Case & Var    &  Learning    & P\_gen                                                  & Q\_gen                                                   & P\_slack            & Q\_slack                                                & V                                                       & $\theta$                                                     \\ 
\midrule

9-bus  & ID   & SL       & \num{0.000018}(\num{0.000020})          & \num{0.000018}(\num{0.000022})          & \num{0.000031}(\num{0.000011})          & \textbf{\num{0.000001}(\num{0.000002})}         & \num{0.000030}(\num{0.000043}) & \num{0.000293}(\num{0.000404}) \\
       &      & +SSL     & \num{0.000008}(\num{0.000011})          & \num{0.000018}(\num{0.000017})          & \textbf{\num{0.000001}(\num{0.000002})} & \num{0.000005}(\num{0.000005})                  & \num{0.000031}(\num{0.000040}) & \num{0.000281}(\num{0.000424}) \\ \cline{2-9} 
       & Line & SL       & \num{0.001764}(\num{0.005105})          & \num{0.018339}(\num{0.038678})          & \num{0.000992}(\num{0.004520})          & \textbf{\num{0.000159}(\num{0.000503})}         & \num{0.000557}(\num{0.001199}) & \num{0.007440}(\num{0.014192}) \\
       &      & +SSL     & \textbf{\num{0.000273}(\num{0.000380})} & \textbf{\num{0.006425}(\num{0.012875})} & \num{0.000405}(\num{0.001104})          & \num{0.000723}(\num{0.001551})                  & \num{0.000431}(\num{0.000718}) & \num{0.007332}(\num{0.014297}) \\ \midrule
30-bus & ID   & SL       & \num{0.000011}(\num{0.000022})          & \num{0.000004}(\num{0.000009})          & \num{0.000001}(\num{0.000002})          & \num{1.123540e-06}(\num{2.276847e-06})          & \num{0.000069}(\num{0.000121}) & \num{0.000199}(\num{0.000790}) \\
       &      & +SSL     & \textbf{\num{0.000008}(\num{0.000019})} & \num{0.000004}(\num{0.000009})          & \num{0.000001}(\num{0.000004})          & \textbf{\num{1.708435e-07}(\num{3.312574e-07})} & \num{0.000071}(\num{0.000137}) & \num{0.000186}(\num{0.000784}) \\ \cline{2-9} 
       & Line & SL       & \num{0.000026}(\num{0.000354})          & \num{0.000015}(\num{0.000140})          & \num{0.000001}(\num{0.000004})          & \num{2.832834e-06}(\num{0.000010})              & \num{0.000127}(\num{0.000466}) & \num{0.000167}(\num{0.000770}) \\
       &      & +SSL     & \num{0.000016}(\num{0.000109})          & \num{0.000055}(\num{0.001353})          & \num{0.000001}(\num{0.000007})          & \textbf{\num{4.066998e-07}(\num{0.000001})}     & \num{0.000127}(\num{0.000461}) & \num{0.000164}(\num{0.000937})
\\ 

      \bottomrule                                 
\end{tabular}
 }
\vspace{2em}
\end{table*}

\textbf{Impact of SSL}
we report in the Table. \ref{tab:pf-ssl} the robustness of powerflow predictions over our 3 grid sizes. We do not consider the scenario of price variation, as the powerflow problem is price-agnostic. Similarly to the insights for the OPF problem, SSL seems to improve the performance of the models with negligible negative impact. SSL improves over SL in six scenarios and marginally degrades the performance in two cases for the reactive power of the slack bus on the 9-bus grid.

\vspace{1em}
\begin{Insight}[boxed title style={colback=pink}]
Learning to optimize the Powerflow problem can also benefit from self-supervised learning, and achieves best robustness against line outage with GAT architectures. 
\end{Insight}
\vspace{1em}


\section{Limitations}

We are releasing our replication package on \url{https://figshare.com/projects/SafePowerGraph_-_NDDS25/212777}. This anonimized repository allows to replicate all the experiments of this work. While our study is the first evaluation of GNN safety for power systems un real-world perturbations, complementary studies could be explored:

\textbf{Distribution grids:} Our work focused on transmission grids, characterized by medium and high voltage and balanced 3-phase nodes. Distribution grids cover low voltages and use a mix of single-phase and three-phase systems and unbalanced loads. Thus the PF and OPF equations for transmission grids are different, and even traditional solvers do not efficiently solve them. Our framework paves the way for fully self-supervised GNN that could solve unbalanced OPF problems without relying on untractable solvers.

\textbf{Other perturbations:} While our work only considered line outages and energy price variations, other manipulation can be considered: The loads (the consumption) of the households can be manipulated sometimes (because of the vulnerability of smart meters), while the line properties (the reactance for example) or the generation capacities are hardly manipulable. 

\textbf{Malicious perturbations:} We only considered random perturbations and did not try to optimize the perturbations to maximize the errors of the GNN. Our study already demonstrated the vulnerability of GNN to random manipulations, our framework should be considered as a minimal benchmark to foster further research on robustness to both random and malicious perturbations.

\textbf{GNN models:} We only considered 3 type of graph layers using undirected graphs and heterogeneous nodes because GNN for OPF achieved best performances with these architectures \cite{ghamizi2024hgnn}. Additional graph and layer architectures can be explored using our framework. 

\vspace{1em}


\section*{Conclusion}

This paper presents SafePowerGraph, a novel, safety-oriented framework designed to address the limitations of existing benchmarks for GNN in PS operations. Our extensive experiments reveal key insights into the robustness and performance of GNN models under realistic, safety-critical scenarios.

First, GNN models are notably vulnerable to price variations across all grid sizes, with bus power flow errors escalating by up to six orders of magnitude for larger grids. Next, GAT architecture consistently outperforms other architectures, demonstrating superior performance and robustness across all scenarios and grid sizes. In addition, integrating SSL with SL enhances model robustness against price variations while maintaining effectiveness on in-distribution load variations and line outages. Lastly, our findings indicate that price variation scenarios lead to significantly more broken power flow and boundary constraints compared to line outage scenarios.

In summary, SafePowerGraph advances the field by providing an open-source, standardized, and robust benchmark that addresses the safety and robustness of GNNs in PS operations. By revealing critical insights and offering a comprehensive evaluation platform, we aim to accelerate progress in developing more resilient and effective GNN models for real-world power systems challenges.

\vspace{1em}


 \section*{Acknowledgment}

This work was supported by FNR CORE project LEAP (17042283).

\clearpage



\bibliographystyle{IEEEtranS}
\bibliography{bib/LIST_LEAP,bib/graph,bib/grid_rob,bib/ml}
\clearpage

\appendices

\section{Problem Definition}
\label{sec:app-A}

\subsection{The Powerflow optimization}
\label{sec:app-A-powerflow}

\textbf{Complex power in AC circuits:}

Electric power transmission is more efficient at high voltages because these higher voltages minimize energy loss due to dissipation in the transmission lines. Power grids typically utilize alternating current (AC) because the voltage of AC can be conveniently changed (from high to low) using transformers. Hence, we will begin by introducing some notation and definitions relevant to AC circuits.

A major feature of AC circuits is that, in contrast to direct current (DC) circuits, the currents and voltages fluctuate over time: their magnitude and direction change periodically. Due to various technical benefits such as reduced losses and fewer disturbances, power generators employ sinusoidal alternating quantities, which can be easily represented using complex numbers.

We will consistently use capital and small letters to denote complex and real-valued quantities, respectively. For instance, let us consider two buses,
$i,j \in \mathcal{N}$, that are directly connected by a transmission line $(i,j)$ . The complex power flowing from bus to bus is denoted by $S_{ij}$ and it can be decomposed into its active $(p_{ij})$ and reactive $(q_{ij})$ components:

\begin{equation}
\label{eq:powerflow}
S_{ij} = p_{ij} + jq_{ij}
\end{equation}

where $j=\sqrt{-1}$ . The complex power flow can be expressed as the product of the complex voltage at bus $i$, $V_i$, and the complex conjugate of the current flowing between the buses $I^*_{ij}$, 

\begin{equation}
\label{eq:complex-powerflow}
S_{ij} = V_i I^*_{ij},
\end{equation}

Transmission lines have power losses due to their resistance $(r_{i,j})$, that indicates the opposition to current flow. In AC circuits, there is an additional dynamic effect due to the line reactance $(x_{ij})$. Unlike resistance, reactance does not result in power loss but causes temporally delays by storing and then returning power to the circuit. The combined effect of resistance and reactance can be expressed through a complex quantity called impedance: $Z_{ij} = r_{i,j} + jx_{ij}$. We also consider the admittance, which is the reciprocal of the impedance: $Y_{ij} = \frac{1}{Z_{ij}}$. Similarly to the impedance, the admittance can be also decomposed into its real, conductance $(g_{ij})$, and imaginary, susceptance $(b_{ij})$, components: $Y_{ij} = g_{ij} ++ jb_{ij}$.

Following Ohm's law, we can write the current as a function of the line voltage drop and the admittance between the two buses:

\begin{equation}
\label{eq:ohm}
I_{ij} = Y_{ij} (V_i-V_j),
\end{equation}

Replacing the above expression for the current in the power flow equation (eq. \ref{eq:complex-powerflow} ), we get

\begin{equation}
\label{eq:complex-line-powerflow}
S_{ij} = Y^*_{ij} V_i V^*_i - Y^*_{ij} V_i V^*_j = Y^*_{ij} (|V_i|^2- V_i V^*_j).
\end{equation}

This power flow equation can be expressed using the admittance components and the polar form of voltage, i.e. $V_i = v_ie^{j \theta_i} = v_i (cos \theta_i + j sin \theta_i)$ (where $v_i$ and $\theta_i$ are the voltage magnitude and angle of bus , respectively):

\begin{equation}
\label{eq:polar-powerflow}
S_{ij} = (g_{ij} - j b_{ij}) (v_i^2 - v_i v_j (cos (\theta_{ij}) + j sin (\theta_{ij}))),
\end{equation}

where we denote with angle difference as $\theta_{ij} = \theta_i-\theta_j$.

We also simplify the conductance and susceptance components with algebraic identities as follows:

\begin{equation}
\label{eq:polar-powerflow-identities}
g_{ij} - j b_{ij} = \frac{g^2_{ij}+ b^2_{ij}}{g_{ij}+ j b_{ij}} = \frac{|Y_{ij}|^2}{Y_{ij}} = \frac{Z_{ij}}{|Z_{ij}|^2} = \frac{r_{i,j} + jx_{ij}}{r^2_{i,j}+x^2_{i,j}},
\end{equation}

hence,the impedance components-based power flow expression becomes:

\begin{equation}
\label{eq:polar-powerflow-final}
S_{ij} = \frac{r_{i,j} + jx_{ij}}{r^2_{i,j}+x^2_{i,j}} (v_i^2 - v_i v_j (cos (\theta_{ij}) + j sin (\theta_{ij}))),
\end{equation}

Finally, the corresponding real equations can be written as:

\begin{equation}
\label{eq:real-powerflow-final}
\resizebox{\linewidth}{!}{$
    \begin{cases}
        p_{ij} = \frac{1}{r^2_{i,j}+x^2_{i,j}} \Bigg[ r_{i,j} (v_i^2 - v_i v_j cos (\theta_{ij})) +  x_{i,j} (v_i v_j sin  (\theta_{ij})) \Bigg] \\
q_{ij} = \frac{1}{r^2_{i,j}+x^2_{i,j}} \Bigg[ x_{i,j} (v_i^2 - v_i v_j cos (\theta_{ij})) +  r_{i,j} (v_i v_j sin  (\theta_{ij})) \Bigg]
    \end{cases}
$}
\end{equation}

\textbf{The bus injection model:}

This last equation models the power flow between two connected buses.  In power flow analysis, the nodes of the whole electrical grid are considered. There are two power flow models based on the type of graph representation employed: the \textbf{bus injection model} (which uses an undirected graph) and the \textbf{branch flow model} (which uses a directed graph).

Our approach is based on undirected graphs, thus we use the bus injection model (BIM), and we introduce it below. 

For any bus $i$, let $\mathcal{N}_i \in \mathcal{N}$ denote the set of buses that are directly linked to bus $i$.

We define as $S_{i,gen}$ the generation power flowing into bus $i$, $S_{i,load}$ the load flowing out of the bus $i$. Then, the net power injection at bus $i$ $S_i =  S_{i,gen} - S_{i,load}$. We denote by $S_{i, trans} = \sum{j \in \mathcal{N}_i}{S_{i,j}}$ the transmitted power flowing between bus $i$ and its adjacent buses.

Tellegen’s theorem states that:

\begin{equation}
\label{eq:bim-basic}
\forall i \in \mathcal{N}, ~ S_{i} = S_{i,gen} - S_{i,load} = S_{i, trans}
\end{equation}

Reusing the complex power flow formulation from \ref{eq:complex-line-powerflow}, the BIM can be expressed as:
 
\begin{align}
\label{eq:bim-complex}
\forall i \in \mathcal{N}, ~ S_{i}  & = S_{i,gen} - S_{i,load} \nonumber \\ 
&= \sum_{j \in \mathcal{N}_i} S_{i,j} \nonumber \\
&=  \sum_{j \in \mathcal{N}_i} Y^*_{ij} (|V_i|^2 - V_i V^*_j)
\end{align}

\textbf{The powerflow problem:}

The BIM equations (eq. \ref{eq:bim-complex}) defines a complex non-linear system with $N = |\mathcal{N}|$ complex equations, and $\{ S_i, V_i\}_{i \in \mathcal{N}}$ complex variables. When using the angular formulation, we construct $2N$ equations using $4N$ real variables $\{p_i, q_i, v_i, \theta_i\}$ with $S_i = p_i + jq_i$.

Based on the variables that are defined, there exist three main types of buses:

\begin{itemize}
    \item The Slack bus (or $V, \theta$) typically serves as a reference bus with predefined voltage angle and magnitude. Slack buses also compensate for any imbalances between generation and demand due to transmission losses. The voltage angle is generally fixed at 0, while the magnitude is maintained at 1.0 per unit.
    \item The load bus (or $PQ$ bus) is a bus with only demand but no power generation. For such buses the active and reactive powers are specified.
    \item The generator bus (or $PV$ bus) is specified with its the active power and voltage magnitude variables. 
\end{itemize}

\textbf{Newton-Raphson Solver}
\label{}
Power flow problems define a non-linear system of equations. There are multiple approaches to solve power flow systems but the most widely used technique is the Newton–Raphson method. Below we demonstrate how it can be applied to the BIM formulation. First, we rearrange eq. \ref{eq:bim-complex}: 

\begin{align}
\label{eq:powerflow}
\forall i \in \mathcal{N}, ~ \mathbf{F}_i = S_{i} - \sum_{j \in \mathcal{N}_i} Y^*_{ij} (|V_i|^2 - V_i V^*_j)
\end{align}

The above set of equations can be expressed simply as $\mathbf{F}(X) = 0$ , where $X$ denotes the real unknown variables and $\mathbf{F}$ represents $2N$ real equations. A solver typically starts from an initial set of unknown variables $X^0$ (ideally close to the solution, or at worse, the steady state of the power grid). The iterative update of $X$ at step $(k+1)$ is then given by:
\begin{equation}
    X^{(k+1)} = X^{(k)} - \mathbf{J}^{-1}(X^{(k)}) \mathbf{F}(X^{(k)}),
\end{equation}
where $\mathbf{J}$ is the Jacobian matrix of partial derivatives of $\mathbf{F}$ with respect to $X$.
This process is repeated until convergence to a threshold defined by the user.

\section{Experimental Protocol}
\label{sec:app-B}

\subsection{GNN outputs and features}
\label{sec:app-B-features}

In our setting, we do not consider load shedding, i.e. reduction of some loads to relieve the network; thus the loads are not controllable.

We report in Table \ref{tab:app-b-output} the outputs for each node type supported in our embedding, and their associated constraints.

We report in Table \ref{tab:app-b-features} the features of each node in the heterogeneous graph. The edges are bidirectional and do not have any features or weights.

\begin{table}[ht]
    
    \caption{Outputs of the GNN models per node type}
    \begin{tabular}{c|c|c|c}
    \toprule
        Node &  Feature & Description & Boundary constraints\\
        \midrule
        Bus & vm\_pu & voltage magnitude [p.u] & [min\_vm\_pu,max\_vm\_pu ]  \\
            & va\_degree & voltage angle [radian] & [-pi, pi] \\
            \midrule
        Generator & p\_mw & Active power [Mw]& [min\_p\_mw,max\_p\_mw ] \\
            & q\_mvar & Reactive power [MVar]& [min\_q\_mvar,max\_q\_mvar ] \\ \midrule
        Slack  & p\_mw & Active power [Mw] & [min\_p\_mw,max\_p\_mw ] \\
            & q\_mvar & Reactive power [MVar]& [min\_q\_mvar,max\_q\_mvar ]  \\
            \bottomrule
    \end{tabular}
    \label{tab:app-b-output}
\end{table}

\begin{table*}[ht]
    
    \caption{Features of the GNN models per node type}
    \begin{tabular}{l|l|l}
    \toprule
        Node &  Feature & Description\\
        \midrule
        Bus & vn\_kv & Initial voltage of the bus [kV] \\
            & min\_vm\_pu & Minimum voltage \\
            & max\_vm\_pu & Maximum voltage \\
            & in\_service & When 0 means that the bus is ignored for PF and OPF calculation [Boolean]\\
            \midrule
        Line& length\_km & Length of the line [km] \\
            & r\_ohm\_per\_km & Resistance [ohm per km] \\
            & x\_ohm\_per\_km &  Reactance [ohm per km] \\
            & c\_nf\_per\_km &  Capacitance [nano Farad per km] \\
            & g\_us\_per\_km & Dielectric conductance [micro Siemens per km] \\
            & max\_i\_ka & Maximal thermal current [kilo Ampere] \\
            & max\_loading\_percent & Maximum loading of the transformer with respect to sn\_mva and its corresponding current at 1.0 p.u. \\
            & in\_service & When 0 means that the line is disconnected for PF and OPF calculation [Boolean]\\
        \midrule
        Transformer  & sn\_mva & Rated power of the load [kVA] \\
            & vn\_hv\_kv & Rated voltage at high voltage bus [kV] \\
            & vn\_Lv\_kv & Rated voltage at low voltage bus [kV] \\
            & vk\_percent & Short circuit voltage [\%] \\
            & vkr\_percent & Real component of short circuit voltage [\%] \\
            & pfe\_kw & Iron losses [kW] \\
            & i0\_percent & 	Open loop losses in [\%] \\
            & shift\_degree & Transformer phase shift angle [deg] \\
            & max\_loading\_percent & 	Open loop losses in [\%] \\
            & in\_service & When 0 means that the transformer is disconnected for PF and OPF calculation [Boolean]\\
            \midrule   
        Load& p\_mw & Active power of the load [MW] \\
            & const\_z\_percent & \% of p\_mw and q\_mvar associated to constant impedance load at rated voltage  \\
            & const\_i\_percent & \% of p\_mw and q\_mvar associated to constant current load at rated voltage \\
            & sn\_mva & Rated power of the load [kVA] \\
            & in\_service & When 0 means that the load is ignored for PF and OPF calculation [Boolean]\\
        \midrule
        Generator &   p\_mw & Initial real power of the load [MW] \\
            & vm\_pu & voltage set point [p.u]  \\
            & sn\_mva & Nominal power of the generator [MVA] \\
            & min\_p\_mw & Minimal active power [Mw] \\
            & min\_p\_mw & Maximal active power [Mw] \\
            & max\_q\_mvar & Minimal reactive power [MVar] \\
            & max\_q\_mvar & Maximal reactive power [MVar] \\
            & in\_service & When 0 means that the generator is ignored for PF and OPF calculation [Boolean]\\
            & cp0\_eur & Offset active power costs [Euro] \\
            & cp1\_eur &  Linear costs per MW [Euro]\\
            & cp2\_eur & Quadratic costs per MW [Euro]\\
            
            \midrule
        Slack    & va\_degree& Angle set point [degree] \\
            & vm\_pu & Voltage set point [p.u]  \\
            & min\_p\_mw & Minimal active power [Mw] \\
            & min\_p\_mw & Maximal active power [Mw] \\
            & max\_q\_mvar & Minimal reactive power [MVar] \\
            & max\_q\_mvar & Maximal reactive power [MVar] \\
            & in\_service & When 0 means that the Slack is ignored for PF and OPF calculation [Boolean]\\
            & cp0\_eur & Offset active power costs [Euro] \\
            & cp1\_eur &  Linear costs per MW [Euro]\\
            & cp2\_eur & Quadratic costs per MW [Euro]\\

             \bottomrule
    \end{tabular}
    \label{tab:app-b-features}
\end{table*}

\section{Additional results}
\label{sec:app-C}

We provide in the following the complementary results for the powerflow problem, and the numerical results for all the experiments

\subsection{Power Flow complementary results}
\label{sec:app-C-pf}

\begin{figure}[!h]
\centering
\begin{subfigure}{0.45\linewidth}
    \includegraphics[clip, trim={0px 0px 250px 0px},width=\textwidth]{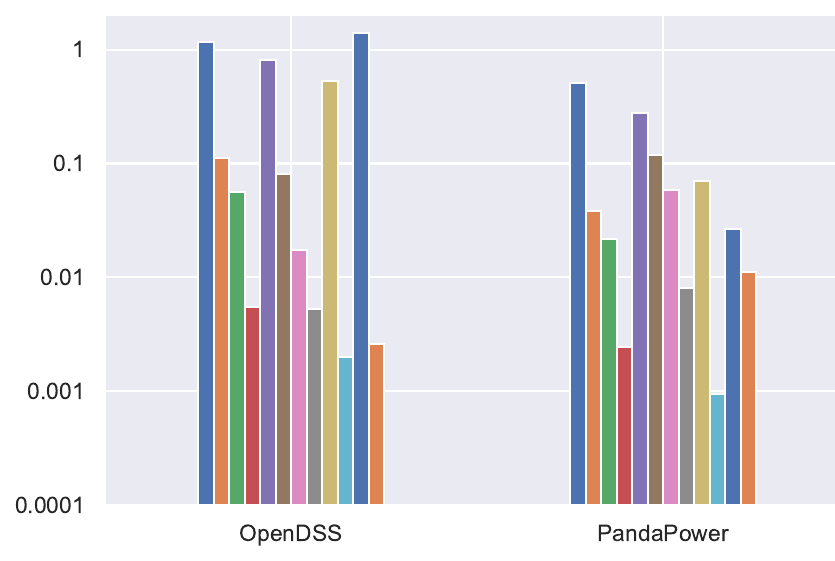}
    \caption{9-bus - ID}
    \label{fig:simulators-pf-case9-id}
\end{subfigure}\begin{subfigure}{0.45\linewidth}
    \includegraphics[clip, trim={0px 0px 250px 0px},width=\textwidth]{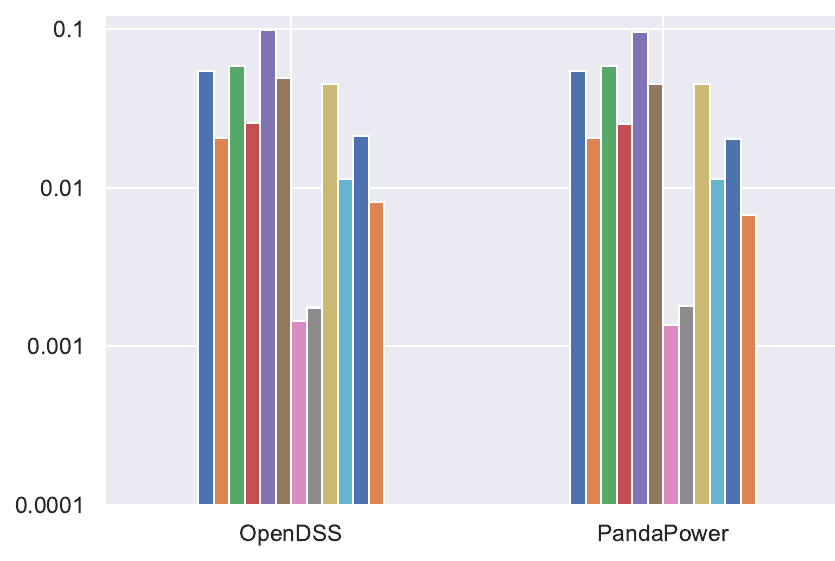}
    \caption{30-bus - ID}
    \label{fig:simulators-pf-case30-id}
\end{subfigure}
\begin{subfigure}{0.45\linewidth}
    \includegraphics[clip, trim={0px 0px 250px 0px},width=\textwidth]{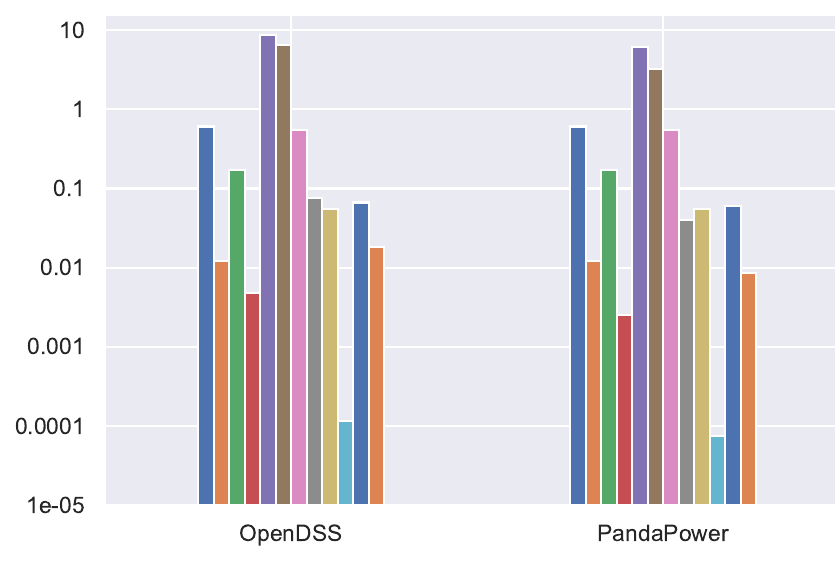}
    \caption{118-bus - ID}
    \label{fig:simulators-pf-case118-id}
\end{subfigure}
\begin{subfigure}{0.45\linewidth}
    \includegraphics[clip, trim={405px 20px 0px 60px},width=\textwidth]{figures/solvers/legend.pdf}
\end{subfigure}
\hfill

\caption{Impact of the simulators on the robustness of the oracle for PowerFlow. Relative MSE between the solutions by MatPower and both Pandapower and OpenDSS}
\label{fig:simulators-pf}
\end{figure}

\textbf{Impact of simulators.}
We study the differences as oracle between the three powerflow simulators MatPower, OpenDSS, and Pandapower on the ID setting in Fig. \ref{fig:simulators-pf}. Our results show much more errors across the solutions found by each solvers. This is to be expected, as the PF problem can have multiple solutions. Thus a reliable benchmark should not consider only the MSE of the oracle solution, but the satisfaction of the power flow and boundary constraints; A feature SafePowerGraph uniquely supports among the frameworks.    

\subsection{Impact of the cost loss}
\label{sec:app-C-cost}

We compare in Table \ref{tab:app-c-cost} different combinations of losses on the I.D scenario: in (1) We use the full loss (supervised loss, ss loss, boundary loss and cost loss), in (2) we discard the cost loss, and in (3) we also discard the SSL loss. 

\begin{table*}[]
    \centering
    
    \caption{Impact of different losses}
    \begin{tabular}{c|c|cccccc}
    \toprule
        {Grid}   & {Scenario}        & $P_g$                        & $Q_g$                        & $P_s$                            & $Q_s$                                                                           & $V$                          & $\theta$                                                                    \\ \midrule
{9-bus}  & {(1) SL+SSL+Cost} & \num{0.000012}                     & \num{0.000011}                     & \num{8.561689e-07}                     & \num{0.000008}                                                                        & \num{0.000044}                     & \num{0.000036}                                                                    \\
{(GAT)}  & {(2) SL+SSL}      & \num{0.000010}                     & \num{0.000014}                     & \num{1.030060e-05}                     & \num{0.000005}                                                                        & \num{0.000048}                     & \num{0.000045}                                                                    \\
         & {(3) SL}           & \num{0.000020} & \num{0.000015} & \num{9.961866e-06} & \num{0.000006}                                                    & \num{0.000046} & \num{0.000137}\\ \midrule
{30-bus} & {(1) SL+SSL+Cost} & \num{0.001361}                & \num{0.001675}                     & \num{1.371326e-03}                     & \num{2.098066e-03}                                                                    & \num{0.002601}                     & \num{0.034110}                                                                    \\
{(SAGE)} & {(2) SL+SSL}      & \num{0.000072}                     & \num{0.000030}                     & \num{4.314451e-05}                     & \num{6.370728e-05}                                                                    & \num{0.001375}                     & \num{0.030541}                                                                    \\ 
         & {(3) SL}           & \num{0.000264} & \num{0.000132} & \num{3.166860e-04} & \num{1.341930e-04} & \num{0.000346} & \num{0.001164}                                               
\\
        \bottomrule
    \end{tabular}
    \label{tab:app-c-cost}
\end{table*}

\subsection{Impact of the weighting strategy}
\label{sec:app-C-weighting}

We evaluate in table the impact of different weighting strategies. Across all loses and rArchs, we report the best value achieved using each weighting strategies between uniform weighting, random weighting, and relative weighting.
The results show that no single strategy stands out in all metrics.

\begin{table*}[]
\centering
    
\caption{Impact of weighting strategies. In bold the best values}
\begin{tabular}{c|c|cccccc}
\toprule
Grid     & Weighting & $P_g$ & $Q_g$ & $P_s$ & $Q_s$ & $V$ & $\theta$ \\ 
\midrule
9-bus    & Uniform   & \num{0.000020} & \num{0.000015} & \num{5.479802e-05} & \textbf{\num{0.000006}} & \num{0.000083} & \num{0.000096} \\
 & Random    & \num{0.000016} & \num{0.000021} & \num{2.774776e-05} & \num{0.000008} & \textbf{\num{0.000044}} & \textbf{\num{0.000036}} \\
         & Relative  & \textbf{\num{0.000012}} & \textbf{\num{0.000011}} & \textbf{\num{8.561689e-07}} & \num{0.000021} & \num{0.000061} & \num{0.000085} \\ 
\midrule
30-bus   & Uniform   & \num{0.000006} & \num{0.000008} & \textbf{\num{9.243440e-07}} & \num{9.214383e-07} & \num{0.000063} & \num{0.000095} \\
     & Random    & \num{0.000005} & \num{0.000005} & \num{2.979039e-06} & \num{1.147172e-06} & \num{0.000070} & \num{0.000125} \\
         & Relative  & \textbf{\num{0.000004}} & \textbf{\num{0.000004}} & \num{4.733350e-06} & \textbf{\num{1.004663e-06}} & \textbf{\num{0.000061}} & \textbf{\num{0.000085}} \\ 
\bottomrule
\end{tabular}
    \label{tab:app-c-weighting}
\end{table*}

\subsection{Detailed numerical results}
\label{sec:app-C-numerical}

In Table \ref{tab:app-c-metrics} we report the best performance achieved for each metric on each grid size and validation set. We report which architecture, loss, weighting strategy and learning rate led to this performance.

Some notation:

- The loss column refers to different combinations (boundary loss is always consdered):
\begin{itemize}
    \item SL: Supervised loss only
    \item SSL: Self-supervised loss only
    \item SL+SSL: Supervised and self supervised losses together
    \item SL+SSL(0,1,1): Starting with supervised loss only, then increasing the weight of the SSL loss from 0 to 1 linearly.
    \item SL+SSL(0,1,2): Starting with supervised loss only, then increasing the weight of the SSL loss from 0 to 1
    \item SL+SSL(0,0.5,1): Starting with supervised loss only, then increasing the weight of the SSL loss from 0 to 0.5 linearly.
    \item SL+SSL+Cost: Supervised, self supervised, and energy cost losses together
\end{itemize}

- LR: refers to the learning rate.

\onecolumn
\begin{longtable}{lllll|llll|rr}
\toprule
Opti                                       & Case    & Training & Validation & Metric    & Loss            & Weighting & LR & Arch & Value     &           \\
                                           &         &          &            &           &                 &           &               &      & mean      & std       \\
\midrule
\endfirsthead
\toprule
Opti                                       & Case    & Training & Validation & Metric    & Loss            & Weighting & LR & Arch & Value     &           \\
                                           &         &          &            &           &                 &           &               &      & mean      & std       \\
\midrule
\endhead
\midrule
\multicolumn{11}{r}{Continued on next page} \\
\midrule
\endfoot
\bottomrule
\endlastfoot
OPF                                        & case9   & ID       & ID         & $P_{gen}$   & SL+SSL(0,0.5,1) & relative   & 0.001         & gcn  & 0.000010  & 0.000013  \\
OPF                                        & case9   & ID       & ID         & $Q_{gen}$   & SL+SSL+Cost     & relative  & 0.001         & gat  & 0.000011  & 0.000014  \\
OPF                                        & case9   & ID       & ID         & $P_{slack}$ & SL+SSL+Cost     & relative  & 0.001         & gat  & 0.000001  & 0.000001  \\
OPF                                        & case9   & ID       & ID         & $Q_{slack}$ & SL+SSL(0,1,1)   & relative   & 0.001         & gat  & 0.000005  & 0.000005  \\
OPF                                        & case9   & ID       & ID         & V         & SL+SSL+Cost     & random    & 0.001         & gat  & 0.000044  & 0.000056  \\
OPF                                        & case9   & ID       & ID         & $\theta$  & SL+SSL+Cost     & random    & 0.001         & gat  & 0.000036  & 0.000043  \\
OPF                                        & case9   & ID       & Price      & $P_{gen}$   & SL+SSL(0,1,2)   & relative   & 0.001         & gat  & 0.094627  & 0.079535  \\
OPF                                        & case9   & ID       & Price      & $Q_{gen}$   & SL+SSL(0,1,2)   & relative   & 0.001         & gat  & 0.003559  & 0.009311  \\
OPF                                        & case9   & ID       & Price      & $P_{slack}$ & SL+SSL          & relative  & 0.001         & gat  & 0.048768  & 0.047070  \\
OPF                                        & case9   & ID       & Price      & $Q_{slack}$ & SL+SSL(0,0.5,1) & relative   & 0.001         & gat  & 0.012945  & 0.014138  \\
OPF                                        & case9   & ID       & Price      & V         & SL              & random    & 0.001         & gat  & 0.002471  & 0.005689  \\
OPF                                        & case9   & ID       & Price      & $\theta$  & SL              & uniform   & 0.001         & gat  & 0.023571  & 0.038714  \\
OPF                                        & case9   & ID       & Line       & $P_{gen}$   & SL+SSL(0,1,2)   & relative   & 0.001         & gcn  & 0.000273  & 0.000380  \\
OPF                                        & case9   & ID       & Line       & $Q_{gen}$   & SL              & relative  & 0.001         & gat  & 0.001006  & 0.001665  \\
OPF                                        & case9   & ID       & Line       & $P_{slack}$ & SL              & random    & 0.001         & gat  & 0.000032  & 0.000037  \\
OPF                                        & case9   & ID       & Line       & $Q_{slack}$ & SL              & relative  & 0.001         & gat  & 0.001992  & 0.005367  \\
OPF                                        & case9   & ID       & Line       & V         & SL              & uniform   & 0.001         & gat  & 0.000641  & 0.001248  \\
OPF                                        & case9   & ID       & Line       & $\theta$  & SSL             & random    & 0.001         & gat  & 0.008625  & 0.015668  \\
OPF                                        & case30  & ID       & ID         & $P_{gen}$   & SL+SSL(0,1,1)   & relative   & 0.001         & gat  & 0.000003  & 0.000009  \\
OPF                                        & case30  & ID       & ID         & $Q_{gen}$   & SL+SSL(0,0.5,1) & relative   & 0.001         & gat  & 0.000002  & 0.000006  \\
OPF                                        & case30  & ID       & ID         & $P_{slack}$ & SL              & uniform   & 0.001         & gat  & 0.000001  & 0.000001  \\
OPF                                        & case30  & ID       & ID         & $Q_{slack}$ & SL+SSL(0,1,1)   & relative   & 0.001         & gat  & 0.000000  & 0.000000  \\
OPF                                        & case30  & ID       & ID         & V         & SL              & relative  & 0.001         & gat  & 0.000061  & 0.000109  \\
OPF                                        & case30  & ID       & ID         & $\theta$  & SL              & relative  & 0.001         & gat  & 0.000085  & 0.000500  \\
OPF                                        & case30  & ID       & Price      & $P_{gen}$   & SL              & uniform   & 0.001         & gcn  & 0.004603  & 0.007646  \\
OPF                                        & case30  & ID       & Price      & $Q_{gen}$   & SL              & random    & 0.001         & gat  & 0.000841  & 0.002431  \\
OPF                                        & case30  & ID       & Price      & $P_{slack}$ & SL              & uniform   & 0.001         & gat  & 0.000449  & 0.000509  \\
OPF                                        & case30  & ID       & Price      & $Q_{slack}$ & SL              & random    & 0.001         & gat  & 0.000294  & 0.001292  \\
OPF                                        & case30  & ID       & Price      & V         & SSL             & random    & 0.001         & gat  & 0.000163  & 0.000425  \\
OPF                                        & case30  & ID       & Price      & $\theta$  & SL              & uniform   & 0.001         & gat  & 0.000666  & 0.001562  \\
OPF                                        & case30  & ID       & Line       & $P_{gen}$   & SL+SSL(0,0.5,1) & relative   & 0.001         & gat  & 0.000024  & 0.000197  \\
OPF                                        & case30  & ID       & Line       & $Q_{gen}$   & SL+SSL(0,0.5,1) & relative   & 0.001         & gcn  & 0.000018  & 0.000047  \\
OPF                                        & case30  & ID       & Line       & $P_{slack}$ & SL+SSL(0,1,1)   & relative   & 0.001         & gat  & 0.000002  & 0.000007  \\
OPF                                        & case30  & ID       & Line       & $Q_{slack}$ & SL+SSL(0,0.5,1) & relative   & 0.001         & gat  & 0.000001  & 0.000001  \\
OPF                                        & case30  & ID       & Line       & V         & SL+SSL(0,0.5,1) & relative   & 0.001         & gat  & 0.000126  & 0.000480  \\
OPF                                        & case30  & ID       & Line       & $\theta$  & SL              & random    & 0.001         & gat  & 0.000156  & 0.000849  \\
OPF                                        & case118 & ID       & ID         & $P_{gen}$   & SL+SSL          & uniform   & 0.001         & gat  & 0.000361  & 0.000908  \\
OPF                                        & case118 & ID       & ID         & $Q_{gen}$   & SL+SSL          & uniform   & 0.001         & gat  & 0.000027  & 0.000091  \\
OPF                                        & case118 & ID       & ID         & $P_{slack}$ & SL              & relative  & 0.001         & gat  & 0.000040  & 0.000021  \\
OPF                                        & case118 & ID       & ID         & $Q_{slack}$ & SL+SSL          & random    & 0.001         & gat  & 0.000021  & 0.000029  \\
OPF                                        & case118 & ID       & ID         & V         & SL              & uniform   & 0.001         & gat  & 0.000208  & 0.000441  \\
OPF                                        & case118 & ID       & ID         & $\theta$  & SL              & relative  & 0.001         & gat  & 0.003209  & 0.008687  \\
OPF                                        & case118 & ID       & Price      & $P_{gen}$   & SL              & random    & 0.001         & gat  & 0.212251  & 0.373310  \\
OPF                                        & case118 & ID       & Price      & $Q_{gen}$   & SL              & random    & 0.001         & gat  & 0.036437  & 0.118089  \\
OPF                                        & case118 & ID       & Price      & $P_{slack}$ & SL              & random    & 0.001         & sage & 0.336550  & 0.291254  \\
OPF                                        & case118 & ID       & Price      & $Q_{slack}$ & SL              & random    & 0.001         & gat  & 0.548407  & 0.601089  \\
OPF                                        & case118 & ID       & Price      & V         & SL              & relative  & 0.001         & sage & 0.003203  & 0.004565  \\
OPF                                        & case118 & ID       & Price      & $\theta$  & SL              & uniform   & 0.001         & gcn  & 0.191129  & 0.452715  \\
OPF                                        & case118 & ID       & Line       & $P_{gen}$   & SL              & random    & 0.001         & gat  & 0.000157  & 0.001153  \\
OPF                                        & case118 & ID       & Line       & $Q_{gen}$   & SL              & random    & 0.001         & gat  & 0.000066  & 0.000620  \\
OPF                                        & case118 & ID       & Line       & $P_{slack}$ & SL+SSL          & uniform   & 0.001         & gat  & 0.000015  & 0.000019  \\
OPF                                        & case118 & ID       & Line       & $Q_{slack}$ & SL+SSL          & random    & 0.001         & gat  & 0.000144  & 0.000618  \\
OPF                                        & case118 & ID       & Line       & V         & SL+SSL          & uniform   & 0.001         & gat  & 0.000250  & 0.000700  \\
OPF                                        & case118 & ID       & Line       & $\theta$  & SL+SSL          & random    & 0.001         & gat  & 0.001851  & 0.011665  \\
\midrule
PF                                         & case9   & ID       & ID         & $P_{gen}$   & SL+SSL          & uniform   & 0.001         & gat  & 0.000008  & 0.000011  \\
PF                                         & case9   & ID       & ID         & $Q_{gen}$   & SL              & uniform   & 0.001         & gat  & 0.000018  & 0.000022  \\
PF                                         & case9   & ID       & ID         & $P_{slack}$ & SL+SSL          & random    & 0.001         & gat  & 0.000001  & 0.000002  \\
PF                                         & case9   & ID       & ID         & $Q_{slack}$ & SL              & random    & 0.001         & gat  & 0.000001  & 0.000002  \\
PF                                         & case9   & ID       & ID         & V         & SL              & random    & 0.001         & gat  & 0.000030  & 0.000043  \\
PF                                         & case9   & ID       & ID         & $\theta$  & SL+SSL          & uniform   & 0.001         & gat  & 0.000281  & 0.000424  \\
PF                                         & case9   & ID       & Price      & $P_{gen}$   & SL+SSL(0,1,2)   & relative   & 0.001         & sage & 0.237373  & 0.237288  \\
PF                                         & case9   & ID       & Price      & $Q_{gen}$   & SL+SSL(0,1,2)   & relative   & 0.001         & sage & 1.108263  & 1.967381  \\
PF                                         & case9   & ID       & Price      & $P_{slack}$ & SL              & uniform   & 0.001         & sage & 0.171705  & 0.264561  \\
PF                                         & case9   & ID       & Price      & $Q_{slack}$ & SL              & uniform   & 0.001         & sage & 2.315716  & 3.025283  \\
PF                                         & case9   & ID       & Price      & V         & SSL             & uniform   & 0.001         & sage & 0.009397  & 0.009332  \\
PF                                         & case9   & ID       & Price      & $\theta$  & SL              & uniform   & 0.001         & sage & 0.556914  & 0.448708  \\
PF                                         & case9   & ID       & Line       & $P_{gen}$   & SL+SSL(0,1,2)   & relative   & 0.001         & sage & 0.000273  & 0.000380  \\
PF                                         & case9   & ID       & Line       & $Q_{gen}$   & SL+SSL(0,1,2)   & relative   & 0.001         & gat  & 0.006425  & 0.012875  \\
PF                                         & case9   & ID       & Line       & $P_{slack}$ & SL+SSL(0,0.5,1) & relative   & 0.001         & gat  & 0.000405  & 0.001104  \\
PF                                         & case9   & ID       & Line       & $Q_{slack}$ & SL              & uniform   & 0.001         & gat  & 0.000159  & 0.000503  \\
PF                                         & case9   & ID       & Line       & V         & SL+SSL(0,1,1)   & relative   & 0.001         & gat  & 0.000431  & 0.000718  \\
PF                                         & case9   & ID       & Line       & $\theta$  & SL+SSL(0,1,1)   & relative   & 0.001         & gat  & 0.007332  & 0.014297  \\
PF                                         & case30  & ID       & ID         & $P_{gen}$   & SL+SSL(0,1,1)   & relative   & 0.001         & gat  & 0.000008  & 0.000019  \\
PF                                         & case30  & ID       & ID         & $Q_{gen}$   & SL+SSL(0,1,2)   & relative   & 0.001         & gat  & 0.000004  & 0.000009  \\
PF                                         & case30  & ID       & ID         & $P_{slack}$ & SL              & uniform   & 0.001         & gat  & 0.000001  & 0.000002  \\
PF                                         & case30  & ID       & ID         & $Q_{slack}$ & SL+SSL(0,1,2)   & relative   & 0.001         & gat  & 0.000000  & 0.000000  \\
PF                                         & case30  & ID       & ID         & V         & SL              & relative  & 0.001         & gat  & 0.000069  & 0.000121  \\
PF                                         & case30  & ID       & ID         & $\theta$  & SL+SSL(0,1,2)   & relative   & 0.001         & gat  & 0.000186  & 0.000784  \\
PF                                         & case30  & ID       & Line       & $P_{gen}$   & SL+SSL(0,1,1)   & relative   & 0.001         & gat  & 0.000016  & 0.000109  \\
PF                                         & case30  & ID       & Line       & $Q_{gen}$   & SL+SSL(0,1,2)   & relative   & 0.001         & gat  & 0.000015  & 0.000140  \\
PF                                         & case30  & ID       & Line       & $P_{slack}$ & SL              & random    & 0.001         & gat  & 0.000001  & 0.000004  \\
PF                                         & case30  & ID       & Line       & $Q_{slack}$ & SL+SSL(0,0.5,1) & relative   & 0.001         & gat  & 0.000000  & 0.000001  \\
PF                                         & case30  & ID       & Line       & V         & SL+SSL          & relative  & 0.001         & gat  & 0.000127  & 0.000461  \\
PF                                         & case30  & ID       & Line       & $\theta$  & SL+SSL(0,1,2)   & relative   & 0.001         & gat  & 0.000164  & 0.000937  \\
PF                                         & case118 & ID       & ID         & $P_{gen}$   & SL              & random    & 0.001         & gat  & 0.000462  & 0.001271  \\
PF                                         & case118 & ID       & ID         & $Q_{gen}$   & SL              & random    & 0.001         & gat  & 0.000045  & 0.000139  \\
PF                                         & case118 & ID       & ID         & $P_{slack}$ & SL              & random    & 0.001         & gat  & 0.000239  & 0.000172  \\
PF                                         & case118 & ID       & ID         & $Q_{slack}$ & SL              & random    & 0.001         & gat  & 0.000165  & 0.000163  \\
PF                                         & case118 & ID       & ID         & V         & SL              & random    & 0.001         & gat  & 0.001920  & 0.003392  \\
PF                                         & case118 & ID       & ID         & $\theta$  & SL              & random    & 0.001         & gat  & 0.010641  & 0.040484  \\
PF                                         & case118 & ID       & Price      & $P_{gen}$   & SL              & random    & 0.001         & gcn  & 0.661156  & 1.429066  \\
PF                                         & case118 & ID       & Price      & $Q_{gen}$   & SL              & random    & 0.001         & gcn  & 2.051154  & 10.525160 \\
PF                                         & case118 & ID       & Price      & $P_{slack}$ & SL              & random    & 0.001         & gcn  & 3.986986  & 3.938960  \\
PF                                         & case118 & ID       & Price      & $Q_{slack}$ & SL              & random    & 0.001         & gcn  & 11.440058 & 10.185546 \\
PF                                         & case118 & ID       & Price      & V         & SL              & random    & 0.001         & gcn  & 0.004513  & 0.005195  \\
PF                                         & case118 & ID       & Price      & $\theta$  & SL              & random    & 0.001         & gcn  & 1.309763  & 1.073374  \\
PF                                         & case118 & ID       & Line       & $P_{gen}$   & SL              & random    & 0.001         & gat  & 0.000283  & 0.001691  \\
PF                                         & case118 & ID       & Line       & $Q_{gen}$   & SL              & random    & 0.001         & gat  & 0.000058  & 0.000610  \\
PF                                         & case118 & ID       & Line       & $P_{slack}$ & SL              & random    & 0.001         & gat  & 0.000070  & 0.000140  \\
PF                                         & case118 & ID       & Line       & $Q_{slack}$ & SL              & random    & 0.001         & gat  & 0.000177  & 0.000337  \\
PF                                         & case118 & ID       & Line       & V         & SL              & random    & 0.001         & gat  & 0.000261  & 0.000478  \\
PF                                         & case118 & ID       & Line       & $\theta$  & SL              & random    & 0.001         & gat  & 0.002354  & 0.010889  \\

\label{tab:app-c-metrics}
\end{longtable}

\twocolumn

\end{document}